\documentclass[11pt]{article}

\usepackage[preprint]{acl}
\usepackage{times}
\usepackage{latexsym}

\usepackage{amsmath}   
\usepackage{amssymb}   

\usepackage{amsmath}  
\usepackage{amssymb}  
\usepackage{amsfonts} 
\usepackage{mathtools} 
\usepackage{bm}       

\usepackage{amsthm}   

\usepackage{booktabs} 
\usepackage{graphicx} 
\usepackage{url}      
\usepackage{hyperref} 

\theoremstyle{plain}

\theoremstyle{definition}

\theoremstyle{remark}

\usepackage{tabularx}

\usepackage{booktabs}
\usepackage{multirow}
\usepackage{graphicx}
\usepackage[table]{xcolor}

\definecolor{bestblue}{HTML}{9BC2E6}   
\definecolor{secondblue}{HTML}{DDEBF7} 

\usepackage{tcolorbox}
\usepackage{enumitem}

\usepackage{algorithm}
\usepackage{amsmath}
\usepackage{amssymb}
\usepackage{bm} 

\usepackage[T1]{fontenc}

\usepackage[utf8]{inputenc}

\usepackage{microtype}

\usepackage{inconsolata}

\usepackage{graphicx}

\title{Self-Correcting RAG: Enhancing Faithfulness via MMKP Context Selection and NLI-Guided MCTS}

\author{
    \textbf{Shijia Xu}$^{\spadesuit}$,
    \textbf{Zhou Wu}$^{\spadesuit, }$\thanks{Corresponding author},
    \textbf{Xiaolong Jia}$^{\heartsuit}$,
    \textbf{Yu Wang}$^{\spadesuit}$,
    \textbf{Kai Liu}$^{\spadesuit, \clubsuit}$,
    \textbf{April Xiaowen Dong}$^{\diamondsuit}$
    \\
    $^{\spadesuit}$Chongqing University, China,
    $^{\heartsuit}$Queen Mary University of London, UK \\
    $^{\clubsuit}$Chongqing Key Laboratory of Big Data Intelligence and Privacy Computing, China \\
    $^{\diamondsuit}$Fangda Partners, China
    \\
    \texttt{\{shijiaxu, ysy\_wang\}@stu.cqu.edu.cn} \\
    \texttt{\{zhouwu, liukai0807\}@cqu.edu.cn,} 
    \texttt{x.jia@qmul.ac.uk}
}

\usepackage{tikz}
\usepackage{pgfplots}
\usepackage{amsmath}  
\usepackage{algorithm} 
\usepackage[noend]{algpseudocode} 
\definecolor{lightblue}{rgb}{0.93, 0.96, 1.0}
\definecolor{lightgray}{gray}{0.9}

\usepackage[utf8]{inputenc}
\usepackage{geometry}

\usepackage{tcolorbox}
\usepackage{xcolor}
\usepackage{amssymb}
\usepackage{pifont}
\usepackage{fontawesome5} 
\usepackage{amsmath}

\definecolor{bg_query}{RGB}{245, 245, 245}       
\definecolor{bg_fail}{RGB}{255, 235, 235}        
\definecolor{bg_success}{RGB}{235, 250, 245}     
\definecolor{border_fail}{RGB}{250, 100, 100}      
\definecolor{border_success}{RGB}{50, 200, 150}    
\definecolor{highlight_text}{RGB}{0, 0, 100}     

\newcommand{\iconSearch}{\faSearch}

\newcommand{\iconFilter}{\faFilter}

\tcbset{
    colback=white,
    colframe=black!60,
    boxrule=0.8pt,
    arc=1.5mm,
    left=1.5mm, right=1.5mm, top=1.5mm, bottom=1.5mm,
    fonttitle=\bfseries\small
}

\definecolor{instblue}{RGB}{218, 232, 252}  
\definecolor{demopink}{RGB}{255, 230, 204}  
\definecolor{inputgreen}{RGB}{213, 232, 212} 

\newtcolorbox{promptbox}[2][]{
  colback=#2,
  colframe=gray!50,
  boxrule=0.5pt,
  arc=2pt,
  left=5pt, right=5pt, top=5pt, bottom=5pt,
  title=\textbf{#1},
  coltitle=black,
  fonttitle=\bfseries\small,
  attach boxed title to top left={xshift=0mm, yshift=0mm},
  boxed title style={empty, frame hidden}
}

\definecolor{myblue}{RGB}{70,130,180}   
\definecolor{myred}{RGB}{220,60,60}     
\definecolor{mygreen}{RGB}{60,160,60}   
\definecolor{mygrey}{RGB}{100,100,100}  

\begin{document}
\maketitle

\begin{abstract}
Retrieval-augmented generation (RAG) substantially extends the knowledge boundary of large language models. However, it still faces two major challenges when handling complex reasoning tasks: low context utilization and frequent hallucinations. To address these issues, we propose Self-Correcting RAG, a unified framework that reformulates retrieval and generation as constrained optimization and path planning. On the input side, we move beyond traditional greedy retrieval and, for the first time, formalize context selection as a multi-dimensional multiple-choice knapsack problem (MMKP), thereby maximizing information density and removing redundancy under a strict token budget. On the output side, we introduce a natural language inference (NLI)-guided Monte Carlo Tree Search (MCTS) mechanism, which leverages test-time compute to dynamically explore reasoning trajectories and validate the faithfulness of generated answers. Experiments on six multi-hop question answering and fact-checking datasets demonstrate that our method significantly improves reasoning accuracy on complex queries while effectively reducing hallucinations, outperforming strong existing baselines.Our code is available at \url{https://github.com/xjiacs/Self-Correcting-RAG}.
\end{abstract}

\section{Introduction}

Large Language Models (LLMs) have shown significant capabilities in reasoning, planning, and tool utilization \citep{bubeck2023sparks, openai2023gpt4, touvron2023llama}. Advanced prompting strategies, such as Chain-of-Thought (CoT) \citep{wei2022chain, kojima2022large}, Least-to-Most prompting \citep{zhou2023least}, and Self-Consistency \citep{wang2023self}, allow models to decompose complex tasks into intermediate steps. Furthermore, recent developments in tool-augmented agents enable models to interact with external environments \citep{mialon2023augmented, schick2023toolformer, qin2024toolllm, shen2024hugginggpt, patil2023gorilla, ruan2023tptu, li2023api}. These advancements suggest a potential for applying LLMs to sophisticated decision-making domains.

\begin{figure}[t!]  
    \centering
    \includegraphics[width=1\linewidth]{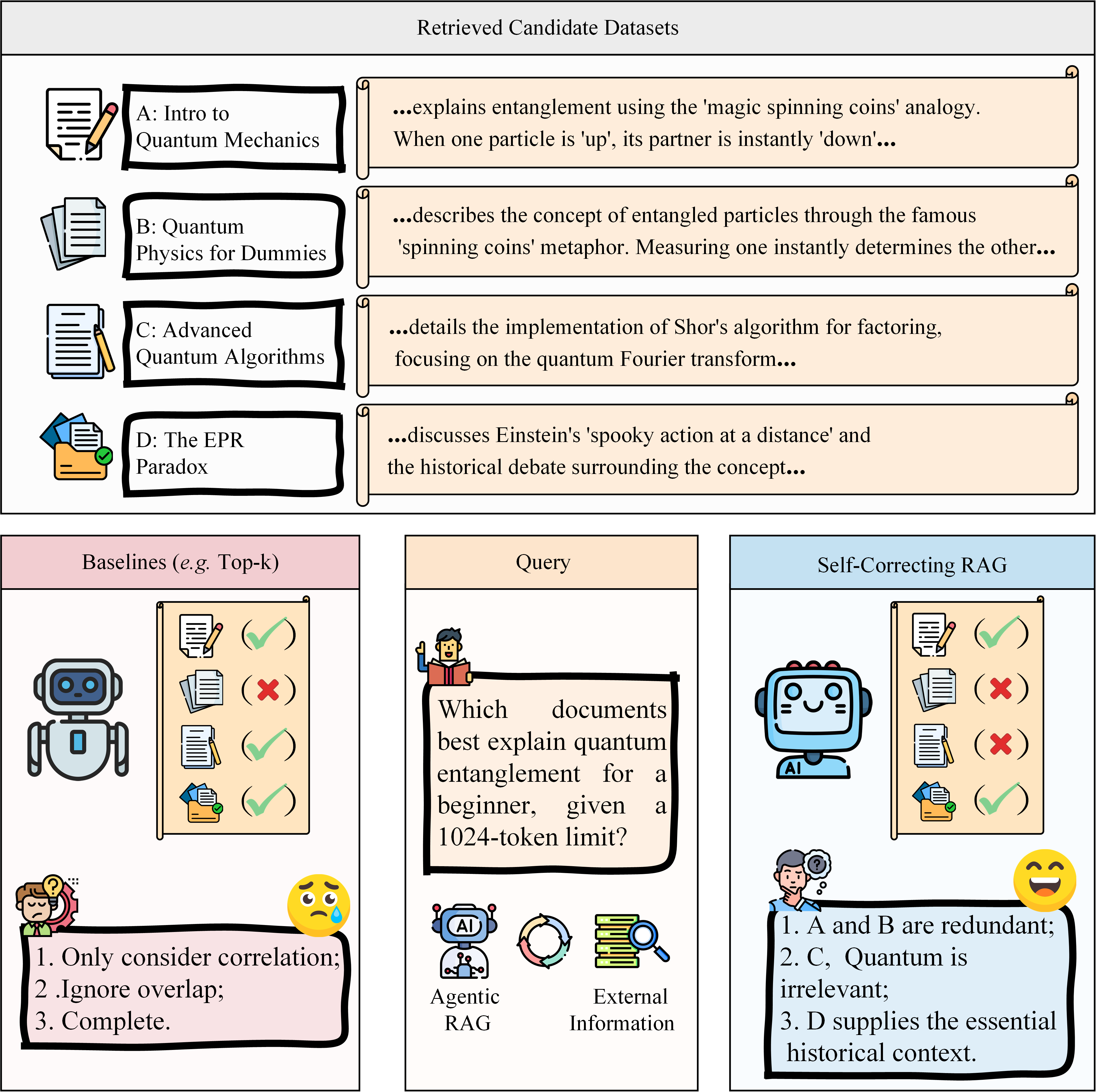}  
    
    \caption{Comparison of document retrieval and selection workflows between the Baseline (Top-k) and the proposed Self-Correcting RAG Framework.}
    \label{fig:framework}
\end{figure}

To support complex reasoning, Retrieval-Augmented Generation (RAG) has emerged as a key strategy. RAG grounds generation in external corpora to mitigate knowledge deficits \citep{lewis2020retrieval, guu2020realm, borgeaud2022improving, izacard2023atlas, ram2023context}. Research in this area has expanded rapidly. Innovations include query rewriting \citep{ma2023query, jagerman2023query}, dense retrieval with pseudo-documents \citep{gao2023precise}, and hierarchical indexing \citep{sarthi2024raptor}. Additionally, self-reflective frameworks have been developed to enhance robustness \citep{asai2024selfrag, yan2024corrective, jiang2023active}. Simultaneously, retrieval components have evolved through advanced embedding models \citep{wang2022text, muennighoff2023mteb, xiao2023cpack} and LLM-based reranking \citep{sun2023chatgpt, ma2023zero, pradeep2023rankvicuna}. These methods provide the necessary evidence to support logical chains.

As shown in Figure~\ref{fig:framework}, applying these capabilities to strictly constrained combinatorial optimization, such as the Multidimensional Multi-choice Knapsack Problem (MMKP), presents distinct challenges. First, LLMs are prone to hallucinations. They may generate plausible but factually incorrect constraints, as noted in recent evaluations \citep{zhang2023siren, huang2023survey, ji2023survey, li2024halueval, min2023factscore, manakul2023selfcheckgpt}. Second, optimization problems involve an exponentially large search space. Greedy token generation cannot guarantee feasibility or global optimality. Errors in early decision steps propagate rapidly, and standard LLMs lack the inherent lookahead mechanisms required for NP-hard problems.

To address the limitations of linear generation, test-time search paradigms organize reasoning into tree or graph structures \citep{yao2024tree, besta2024graph, shinn2024reflexion, zelikman2022star, long2023large}. Approaches such as Language Agent Tree Search (LATS) integrate Monte Carlo Tree Search (MCTS) with language agents \citep{zhou2024language, wang2024rest, hao2023reasoning}. Concurrently, LLMs are increasingly utilized as heuristics or optimizers for combinatorial tasks \citep{yang2024large, ye2024reevo, romeraparedes2024mathematical, guo2023connecting, liu2023llm}. These works indicate that a robust solver must combine retrieval-based evidence with structured exploration.

In this paper, we propose a unified framework that synergizes RAG with MCTS to address the MMKP. Our approach grounds local feasibility checks in traceable evidence while organizing the decision process into a backtrackable tree. The main contributions of this work are summarized as follows:

\begin{itemize}
    \item We introduce a MMKP-based Context Selector that models document selection as a constrained knapsack problem. This method maximizes information density under token budgets while minimizing redundancy, outperforming greedy ranking strategies.
    \item We develop an NLI-Guided MCTS Generator that utilizes test-time compute to explore reasoning paths. By using Natural Language Inference as a reward model, we penalize contradictions and ensure the generated answers are faithful to the retrieved context.
    \item We achieve strong performance across six diverse datasets, including multi-hop QA and fact verification benchmarks. Our framework significantly reduces hallucinations and improves reasoning accuracy compared to strong agentic baselines.
\end{itemize}

\section{Related Work}

\subsection{Retrieval-Augmented Generation and Reasoning}
Prompting strategies have significantly enhanced the reasoning capabilities of Large Language Models (LLMs). Chain-of-Thought (CoT) \citep{wei2022chain, kojima2022large} and ensemble methods like Self-Consistency \citep{wang2023self} established baselines for intermediate reasoning. Beyond static generation, agentic frameworks such as ReAct \citep{yao2023react} and Toolformer \citep{schick2023toolformer} empower models to execute actions and utilizing APIs \citep{qin2024toolllm, patil2023gorilla}. However, these agents primarily operate in open-ended environments rather than strictly constrained decision spaces.

A critical challenge in these reasoning tasks is hallucination, where models generate fact-conflicting content \citep{ji2023survey, zhang2023siren}. Retrieval-Augmented Generation (RAG) addresses this by grounding outputs in external corpora \citep{lewis2020retrieval, guu2020realm}. Recent advancements focus on robustness, utilizing hierarchical indexing \citep{sarthi2024raptor} and corrective mechanisms like Self-RAG \citep{asai2024selfrag} and CRAG \citep{yan2024corrective}. Concurrently, retrieval components have improved via instruction-tuned embeddings \citep{wang2022text, muennighoff2023mteb} and listwise reranking \citep{sun2023chatgpt, pradeep2023rankvicuna}, ensuring relevant context is prioritized to mitigate knowledge deficits.

\subsection{LLMs for Combinatorial Optimization}
Research increasingly explores LLMs as optimizers or heuristics for hard problems. OPRO \citep{yang2024large} demonstrates that LLMs can iteratively improve solutions via natural language prompts, while ReEvo \citep{ye2024reevo} leverages models to evolve heuristics. FunSearch \citep{romeraparedes2024mathematical} further illustrates the potential of pairing LLMs with evolutionary strategies to discover mathematical constructions.

Specific to Combinatorial Optimization (CO), prior work has applied LLMs to routing tasks like the Traveling Salesperson Problem (TSP) \citep{liu2023llm, guo2023connecting, ahn2022can}. However, strictly constrained problems like MMKP remain under-explored compared to routing. This is largely due to the difficulty of maintaining feasibility; a single hallucinated constraint can invalidate a solution \citep{li2024halueval}. Evaluation benchmarks such as FActScore \citep{min2023factscore} and SelfCheckGPT \citep{manakul2023selfcheckgpt} highlight the fragility of LLMs in adhering to strict conditions. Our framework addresses this by aligning optimization steps with verifiable evidence, treating valid constraint satisfaction as a prerequisite \citep{turpin2024language}.

\subsection{Tree Search and Test-Time Computation}
To overcome the limitations of linear decoding, recent methods organize reasoning into non-linear structures. Approaches like Tree of Thoughts (ToT) \citep{yao2024tree} and Graph of Thoughts (GoT) \citep{besta2024graph} generalize CoT by maintaining multiple reasoning paths. Reflexion \citep{shinn2024reflexion} adds a verbal reinforcement learning layer through self-reflection to refine outputs.

Monte Carlo Tree Search (MCTS) has emerged as a powerful mechanism for complex decision-making within this landscape. Language Agent Tree Search (LATS) \citep{zhou2024language} unifies planning, acting, and reasoning within an MCTS framework. Similarly, reasoning-via-planning methods \citep{hao2023reasoning} demonstrate the efficacy of MCTS in structured tasks. This test-time search paradigm allows models to trade inference compute for solution quality. By systematically exploring the decision space, these methods provide the lookahead capabilities required for optimization, which standard greedy generation lacks.

\section{Methodology}
\label{sec:methodology}

\begin{figure*}[t]
    \centering 
    \includegraphics[width=0.87\textwidth]{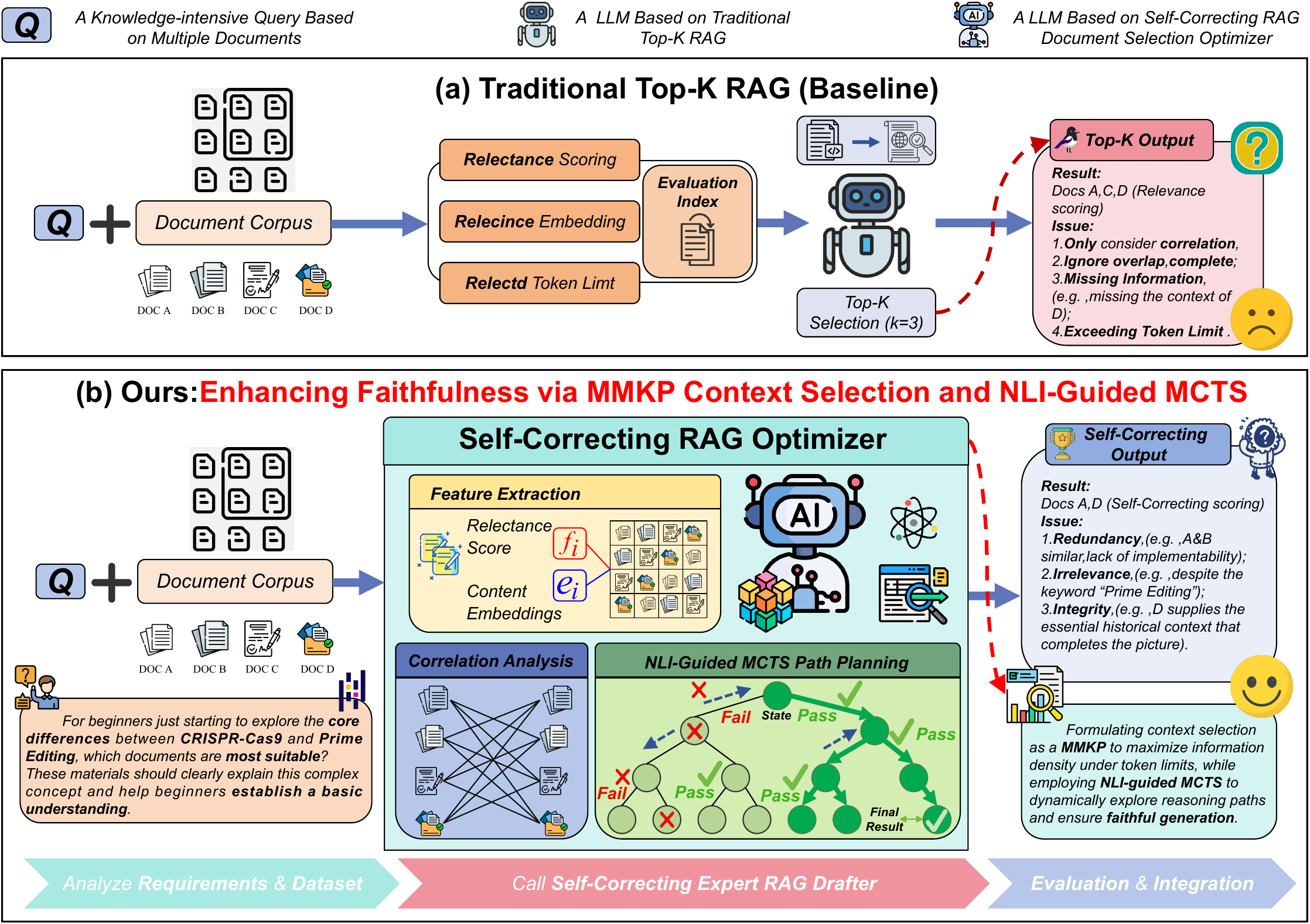}
    \caption{Illustration of the comparison between traditional retrieval paradigms and our proposed framework.(a) The Baseline Traditional Top-K RAG relies primarily on simple relevance scoring and embeddings to select top-ranked documents within a token limit.(b) In contrast, our proposed Self-Correcting RAG Optimizer models document selection as a combinatorial optimization problem. It integrates feature extraction and a dedicated Self-Correcting Optimization Engine (employing MMKP and NLI-guided mechanisms) to efficiently select the best draft.}
 \label{fig:selfrag}
\end{figure*}

We present a rigorous theoretical framework that bifurcates the RAG optimization problem into two distinct phases, as illustrated in Figure~\ref{fig:selfrag}. It comprises pre-generation context optimization, modeled as a Multiple-Choice Multidimensional Knapsack Problem (MMKP), and inference-time reasoning, modeled as a logic-guided Monte Carlo Tree Search (MCTS) process.

\subsection{Phase I: Optimal Context Selection via MMKP}
\label{sec:mmkp_formulation}

Let $\mathcal{U}$ be the universe of retrieved document chunks for a given query $q$. Conventional methods treat $\mathcal{U}$ as a flat list, selecting top-$k$ by relevance score $S_{rel}(q, d)$. We argue this is suboptimal due to high inter-document redundancy.

We formally recast context selection as selecting a subset $\mathcal{S} \subseteq \mathcal{U}$ to maximize information density under multidimensional constraints.

\subsubsection{Semantic Grouping and Problem Definition}
First, we induce a partition on $\mathcal{U}$ via semantic clustering to enforce diversity. Let $\Phi(d) \in \mathbb{R}^D$ be the dense embedding of document $d$. We define the groups $\mathcal{G} = \{G_1, G_2, \dots, G_m\}$ such that for any $G_i$, $\forall d_a, d_b \in G_i$, the cosine similarity $\cos(\Phi(d_a), \Phi(d_b)) \ge \tau$, where $\tau$ is a similarity threshold. This implies that items within $G_i$ are mutually exclusive candidates for the context window (i.e., choosing multiple provides diminishing returns).

Let $d_{ij}$ denote the $j$-th document in group $G_i$, and we introduce a binary decision variable $x_{ij} \in \{0,1\}$ where $x_{ij}=1$ iff $d_{ij}$ is selected. The full definition of the multidimensional cost vectors $\mathbf{w}_{ij}$ (token consumption and redundancy penalty) and the fused utility $v_{ij}$ are provided in Appendix~\ref{app:mmkp_defs}, with the MMKP objective unchanged.

\subsubsection{The MMKP Optimization Objective}
The objective is to maximize total utility $Z$ subject to the capacity vector $\mathbf{C} = [C_{token}, C_{red}]^T$:

\begin{equation}
\label{eq:mmkp_objective}
\begin{aligned}
    \text{maximize} \quad & Z(\mathbf{x}) = \sum_{i=1}^{m} \sum_{j=1}^{|G_i|} v_{ij} x_{ij} \\
    \text{subject to} \quad & \sum_{i=1}^{m} \sum_{j=1}^{|G_i|} \mathbf{w}_{ij} x_{ij} \preceq \mathbf{C} \quad  \\
    & \sum_{j=1}^{|G_i|} x_{ij} \le 1, \quad \forall i \in \{1, \dots, m\} \\
    & x_{ij} \in \{0, 1\}
\end{aligned}
\end{equation}

The multiple-choice constraint ($\sum x_{ij} \le 1$) enforces that we select at most one representative from each semantic cluster, thereby maximizing information coverage. The NP-hardness of this formulation, proven via a reduction, is detailed in Appendix~\ref{app:proof_np_hard}.

\subsection{Phase II: Inference-Time Reasoning via NLI-Guided MCTS}
\label{sec:mcts_algo}

While MMKP optimizes the input context, it does not guarantee that the generated answer is faithful. To address hallucinations, we introduce a Test-Time Compute strategy modeled as a Markov Decision Process (MDP) solved via Monte Carlo Tree Search.

\subsubsection{MDP Formulation}
We define the Markov Decision Process as the tuple $(\mathcal{S}, \mathcal{A}, \mathcal{P}, \mathcal{R})$.

\noindent \textbf{State Space $\mathcal{S}$.} A state $s_t = (q, \mathcal{D}_{ctx}, y_{1:t-1})$ consists of the query, the currently selected context documents, and the partial answer generated so far.

\noindent \textbf{Action Space $\mathcal{A}$.} At each step, the policy network (LLM) selects one of two actions. A generative action $a_{gen}$ samples a continuation from $P_{LLM}(y_t \mid s_t)$. An augmentative action $a_{aug}$ triggers a retrieval call based on the uncertainty of $y_{1:t-1}$ to obtain $\mathcal{D}_{new}$ and update the context as $\mathcal{D}_{ctx}' = \mathcal{D}_{ctx} \cup \mathcal{D}_{new}$.

\noindent \textbf{Transition $\mathcal{P}$.} The transition is deterministic for $a_{aug}$ and stochastic for $a_{gen}$, governed by the LLM logits.





\subsubsection{The NLI Reward Function}
We define a dense reward function $\mathcal{R}(s)$ that evaluates the logical entailment between the generated answer sentences and the retrieved evidence.
Let the generated answer $y$ be split into sentences $\{u_1, \dots, u_L\}$. Let $\mathcal{E} = \{e_1, \dots, e_K\}$ be the top-$K$ evidence snippets extracted from $\mathcal{D}_{ctx}$.

We employ a Natural Language Inference (NLI) model $\Theta_{NLI}(e, u) \to [0,1]^3$ mapping to probabilities $\{P_{ent}, P_{neu}, P_{con}\}$. The reward is computed as:
\begin{equation}
\label{eq:nli_reward}
    R(y, \mathcal{D}_{ctx}) = \frac{1}{L} \sum_{l=1}^{L} \max_{e \in \mathcal{E}} \left[ \mathbf{W}^T \cdot \Theta_{NLI}(e, u_l) \right]
\end{equation}
where $\mathbf{W} = [w_{ent}, w_{neu}, w_{con}]^T$ is the weight vector. We explicitly set a severe penalty for contradictions ($w_{con} \ll 0$) to prune hallucinated branches.Details of the UCT \& PUCT selection, expansion, rollout, and backpropagation procedure are provided in Appendix~\ref{app:mcts_details}. Furthermore, a convergence \& consistency justification is given in Appendix~\ref{thm:mcts_consistency}.

\begin{table*}[h]
  \centering
  \begin{tabular}{lcccccc}
    \toprule
    & NQ & PopQA & MuSiQue & 2Wiki & HotpotQA & MultiHop-RAG \\
    \midrule
    Num of queries & 1,000 & 1,000 & 1,000 & 1,000 & 1,000 & 2,556 \\
    Num of passages & 9,633 & 8,676 & 11,656 & 6,119 & 9,811 & 609 \\
    \bottomrule
  \end{tabular}
  \smallskip
\caption{Dataset statistics}
\label{sec:data}
\end{table*}

\subsection{Approximation Algorithms for MMKP}
\label{sec:approx_mmkp}

Solving Eq.~\ref{eq:mmkp_objective} exactly is computationally prohibitive ($O(m \cdot 2^{|G_{max}|})$). We provide two solutions: a theoretical FPTAS for the single-dimensional case, with its proof details provided in Appendix~\ref{app:fptas_proof}; and a practical heuristic for the multi-dimensional case based on Pareto-pruned DP, whose details are provided in Appendix~\ref{app:pareto_pruning_impl}. For the practical implementation where $D=2$, we use a Dynamic Programming approach with Pareto pruning to control state growth.


\section{Experiments}
\label{sec:experiments}



\subsection{Datasets}
To rigorously evaluate our Self-Correcting RAG, we conduct extensive experiments across six challenging datasets of three tasks, ranging from single-hop retrieval to complex multi-step reasoning. Dataset statistics are summarized in Table~\ref{sec:data}, with comprehensive details provided in Appendix~\ref{app:datasets}.

\noindent \textbf{Simple QA.} 
We evaluate open-domain question answering using \textbf{NQ}, a challenging subset of the Natural Questions benchmark~\cite{kwiatkowski2019natural}, designed to evaluate open-domain question answering under combined retrieval and reasoning demands. We then evaluate performance on long-tail knowledge using \textbf{PopQA}~\cite{mallen2023popqa}, which targets rare entities and infrequent facts that are typically missing from parametric memory and are known to trigger hallucinations in standard language models.

\noindent \textbf{Multi-Hop QA.} We evaluate models on three representative datasets to assess their complex information aggregation capability. We adopt \textbf{MuSiQue}~\cite{trivedi2022musique}, which is designed to avoid shortcut learning. \textbf{2WikiMultiHopQA}~\cite{ho20202wikimultihopqa} is included to evaluate structured reasoning abilities, as it involves reasoning chains with entity relations and comparative logic. We use \textbf{HotpotQA}~\cite{yang2018hotpotqa} in the distractor setting, which requires models to bridge information across two distinct documents to derive correct answers. 

\noindent \textbf{Multi-Doc QA.} Real-world retrieval is often imperfect. We include \textbf{MultiHop-RAG}~\cite{tang2024multihoprag}, a benchmark specifically constructed to evaluate resilience against noisy, irrelevant, and misleading context. This dataset tests the system's ability to filter out red herring documents that share lexical overlap with the query but contain no answer-bearing evidence.


\subsection{Baselines}
We compare our approach with three categories of retrieval-augmented generation methods, ranging from standard pipelines to advanced agentic frameworks.

\noindent \textbf{Standard RAG Baselines.} We utilize \textbf{Naive RAG} as a primary baseline, following the Retrieve then Generate paradigm with BGE-Large embeddings and top-$k$ truncation. To evaluate query optimization, we include \textbf{HyDE}~\cite{gao2023precise}, which generates hypothetical documents to bridge the semantic gap, and \textbf{RRR}~\cite{ma2023query}, which employs an LLM to rewrite input queries for better alignment with the corpus.

\noindent \textbf{Advanced Selection \& Reranking.} These methods focus on optimizing the context window. \textbf{Filco}~\cite{wu2024filco} leverages lexical and semantic signals to filter out irrelevant chunks post-retrieval, while \textbf{RECOMP}~\cite{xu2024recomp} maximizes information density by compressing retrieved documents into concise textual summaries to reduce noise and context length. \textbf{LongLLMLingua}~\cite{wang2023longllmlingua} adopts a hierarchical text compression strategy, leveraging pre-trained language models to iteratively prune redundant content.

\noindent \textbf{Iterative \& Agentic RAG.} We benchmark against state-of-the-art dynamic frameworks. \textbf{IRCoT}~\cite{liu2023cotrag} guides retrieval via step-by-step reasoning. \textbf{Self-RAG}~\cite{asai2024selfrag} trains the generator to output reflection tokens for self-critique. \textbf{CRAG}~\cite{yan2024corrective} incorporates a lightweight evaluator to trigger web searches when retrieval is ambiguous. \textbf{DRAG}~\cite{zhang2024drag} enables dynamic retrieval adjustment based on intermediate results.

\subsection{Metrics}
We adopt a multi-dimensional evaluation protocol. For generation quality, we report Exact Match (EM) and F1 Score. For retrieval quality, we compute Recall@5 to evaluate the MMKP selection. To assess faithfulness, we report Citation Precision and Contradiction Rate, verified by an NLI model (RoBERTa-large-mnli) to ensure generated claims are entailed by their cited evidence.

\subsection{Implementation Details}
\label{sec:implementation}

For Self-Correcting RAG, we employ Qwen2.5-7B-Instruct as the backbone generator. We use BAAI/bge-small-en-v1.5 as the dense retriever and BM25 for sparse retrieval, fusing results via Reciprocal Rank Fusion (RRF). The detailed prompts are shown in Appendix~\ref{app:prompts}. For the MMKP, we operate with specified token and redundancy budgets, and utilize a dynamic programming solver with Pareto pruning. Regarding MCTS, we use RoBERTa-large-mnli to strictly penalize contradictions and neutral outputs. All experiments were conducted on a cluster of $8 \times$ NVIDIA A100 (80GB) GPUs. More implementation and hyperparameter details can be found in Appendix~\ref{sec:appendix_implementation}.

\begin{table*}[t!]
\centering 

\begin{minipage}{\textwidth}

\centering
\resizebox{0.98\textwidth}{!}{%
\begin{tabular}{lcccccccccccccc}
\toprule
\multirow{3}{*}{\textbf{Method}} 
& \multicolumn{4}{c}{\textbf{Simple QA}} 
& \multicolumn{6}{c}{\textbf{Multi-Hop QA}} 
& \multicolumn{2}{c}{\textbf{Multi-Doc QA}} 
& \multicolumn{2}{c}{\multirow{2}{*}{\textbf{Avg}}} \\
\cmidrule(lr){2-5} \cmidrule(lr){6-11} \cmidrule(lr){12-13} 

& \multicolumn{2}{c}{NQ} & \multicolumn{2}{c}{PopQA} 
& \multicolumn{2}{c}{MuSiQue} & \multicolumn{2}{c}{2Wiki} & \multicolumn{2}{c}{HotpotQA} 
& \multicolumn{2}{c}{MultiHop-RAG} 
& \multicolumn{2}{c}{} \\
\cmidrule(lr){2-3} \cmidrule(lr){4-5} \cmidrule(lr){6-7} \cmidrule(lr){8-9} \cmidrule(lr){10-11} \cmidrule(lr){12-13} \cmidrule(lr){14-15} 
 & EM & F1 & EM & F1 & EM & F1 & EM & F1 & EM & F1 & EM & F1 & EM & F1 \\
\midrule

\rowcolor{lightgray} \multicolumn{15}{c}{\textbf{\textit{Standard RAG Baselines}}} \\

Naive & 39.4 & 52.1 & 37.6 & 48.4 & 14.6 & 25.8 & 14.8 & 24.3 & 25.8 & 35.8 & 22.4 & 30.3 & 25.8 & 36.1 \\
HyDE & 44.5 & 53.2 & 39.8 & 49.5 & 18.2 & 27.5 & 23.5 & 31.8 & 33.6 & 41.2 & 24.5 & 32.1 & 30.7 & 39.2 \\
RRR & 45.1 & 53.8 & 40.2 & 50.1 & 20.5 & 29.5 & 26.1 & 35.2 & 36.5 & 44.8 & 25.1 & 32.8 & 32.3 & 41.0 \\

\rowcolor{lightgray} \multicolumn{15}{c}{\textbf{\textit{Advanced Selection \& Reranking}}} \\

RAG + MMR & 46.2 & 54.5 & 41.5 & 50.8 & 19.8 & 28.4 & 27.8 & 36.5 & 38.2 & 46.5 & \underline{32.1} & \underline{40.5} & 34.3 & 42.9 \\
Filco & 45.8 & 54.2 & 42.1 & 51.2 & 19.5 & 28.1 & 27.2 & 36.1 & 37.8 & 46.2 & 29.8 & 37.2 & 33.7 & 42.2 \\
RECOMP & \underline{47.1} & \underline{55.4} & 41.8 & 50.5 & \underline{21.5} & \underline{30.2} & 29.5 & 38.5 & 39.5 & 48.1 & 29.2 & 36.8 & \underline{34.8} & \underline{43.3} \\
LongLLMLingua & 43.8 & 52.6 & 40.5 & 49.8 & 17.8 & 26.5 & 25.5 & 34.2 & 35.5 & 43.5 & 26.2 & 33.5 & 31.6 & 40.0 \\

\rowcolor{lightgray} \multicolumn{15}{c}{\textbf{\textit{Iterative \& Agentic RAG}}} \\

IRCoT & 42.5 & 51.8 & 40.8 & 49.2 & 18.5 & 27.8 & \underline{30.1} & \underline{39.2} & 40.5 & 48.2 & 27.5 & 35.2 & 33.3 & 41.9 \\
Self-RAG & 43.2 & 52.5 & 38.5 & 46.8 & 20.2 & 29.1 & 22.5 & 31.5 & 40.8 & \textbf{51.2}$^{\dagger}$ & 28.1 & 36.2 & 32.2 & 41.2 \\
CRAG  & 40.5 & 50.1 & \textbf{45.5}$^{\dagger}$ & \textbf{54.5}$^{\dagger}$ & 18.2 & 27.8 & 27.5 & 37.2 & \textbf{42.5}$^{\dagger}$ & \underline{50.1} & 31.5 & 40.2 & 34.3 & \underline{43.3} \\
DRAG & 36.8 & 50.4 & 38.6 & 46.5 & 20.4 & 28.1 & 28.8 & 37.0 & 30.8 & 41.7 & 29.3 & 30.2 & 30.8 & 39.0 \\
\midrule
\rowcolor{lightblue}
\textbf{Self-Correcting RAG} & \textbf{48.4}$^{\dagger}$ & \textbf{56.2}$^{\dagger}$ & \underline{43.2} & \underline{51.9} & \textbf{22.7}$^{\dagger}$ & \textbf{31.9}$^{\dagger}$ & \textbf{31.2}$^{\dagger}$ & \textbf{40.6}$^{\dagger}$ & \underline{41.8} & 49.1 & \textbf{35.3}$^{\dagger}$ & \textbf{44.8}$^{\dagger}$ & \textbf{37.1}$^{\dagger}$ & \textbf{45.8}$^{\dagger}$ \\ 
\bottomrule
\end{tabular}%
}

\caption{The performance comparison across Simple QA, Multi-Hop QA, and Multi-Doc QA benchmarks using Exact Match (EM) and F1 scores. The datasets include NQ and PopQA for simple queries, MuSiQue, 2Wiki, and HotpotQA for multi-hop reasoning, and MultiHop-RAG for multi-document contexts. The best performance is bolded with $^{\dagger}$ and the second best is underlined.}
\label{tab:comparison}
\end{minipage}%
\hfill 


\begin{minipage}{0.8\textwidth}

\centering
\resizebox{\textwidth}{!}{%
\begin{tabular}{lccccccc}
\toprule
\multirow{2}{*}{\textbf{Method}}
& \multicolumn{2}{c}{\textbf{Simple QA}}
& \multicolumn{3}{c}{\textbf{Multi-Hop QA}}
& \multicolumn{1}{c}{\textbf{Multi-Doc QA}}
& \multirow{2}{*}{\textbf{Avg}} \\
\cmidrule(lr){2-3} \cmidrule(lr){4-6} \cmidrule(lr){7-7}
 & NQ & PopQA & MuSiQue & 2Wiki & HotpotQA & MultiHop-RAG & \\
\midrule
\rowcolor{lightgray} \multicolumn{8}{c}{\textbf{\textit{Standard Baselines}}} \\
Naive & 56.1 & 35.7 & 43.5 & 65.3 & 74.8 & 22.4 & 49.6 \\
HyDE & 58.6 & 43.2 & 46.6 & 67.5 & 75.3 & 25.1 & 52.7 \\
RRR & 63.4 & 49.4 & 49.1 & 67.9 & 78.9 & 28.6 & 56.2 \\
\rowcolor{lightgray} \multicolumn{8}{c}{\textbf{\textit{Advanced Selection \& Reranking}}} \\
RAG + MMR & 68.3 & 55.6 & 63.6 & 74.8 & 85.1 & 32.2 & 63.3 \\
Filco & 70.1 & 58.2 & 65.9 & 76.0 & 88.4 & 34.8 & 65.6 \\
RECOMP & \underline{71.5} & 60.1 & 69.7 & 80.5 & 90.2 & \underline{38.5} & 68.4 \\
LongLLMLingua & 67.5 & 53.8 & 61.2 & 73.5 & 85.0 & 31.5 & 62.1 \\
\rowcolor{lightgray} \multicolumn{8}{c}{\textbf{\textit{Iterative \& Agentic RAG}}} \\
IRCoT & 65.2 & 52.7 & 57.8 & 70.2 & 86.9 & 30.3 & 60.5 \\
Self-RAG & 67.8 & 54.5 & 55.9 & 79.1 & 87.7 & 33.6 & 63.1 \\
CRAG & 69.5 & 58.7 & \textbf{74.8} & 82.5 & \underline{91.5} & 35.4 & \underline{68.7} \\
DRAG & 64.4 & \underline{61.8} & 60.2 & \underline{84.4} & 88.3 & 31.1 & 65.0 \\
\midrule
\rowcolor{lightblue}
\textbf{Self-Correcting RAG} & \textbf{72.8} & \textbf{65.5} & \underline{72.5} & \textbf{86.9} & \textbf{93.6} & \textbf{40.4} & \textbf{72.0} \\
\bottomrule
\end{tabular}%
}
\caption{Retrieval performance (passage recall@5) on RAG benchmarks. While CRAG achieves the best performance on MuSiQue, Self-Correcting RAG demonstrates superior consistency, achieving the highest recall on 5 out of 6 datasets.}
\label{tab:retrieval_performance}
\end{minipage}

\end{table*}

\section{Results}
\label{sec:results}
We now present our main QA and retrieval experimental results, where the QA process uses retrieved results as its context. More detailed experimental results are presented in Appendix~\ref{app:Results}.

\subsection{Generation Performance}
Table~\ref{tab:comparison} summarizes the Exact Match (EM) and F1 scores across six diverse datasets. Our method achieves the highest average performance among all evaluated models. Specifically, it surpasses the strongest baselines with an average EM of 37.1 and an average F1 score of 45.8.

\paragraph{Complex Reasoning Tasks.}
The advantages of our approach are most pronounced in complex multi-hop reasoning scenarios, such as MuSiQue and 2WikiMultiHopQA. These tasks require the model to aggregate disparate pieces of information from multiple documents. On the MuSiQue dataset, our Self-Correcting RAG outperforms the previous state-of-the-art model, CRAG, by a substantial margin. While CRAG achieves an EM of 18.2, our method reaches 22.7, representing a 4.5\% absolute improvement. This significant gain indicates that the NLI-guided MCTS generator effectively navigates complex reasoning paths. It succeeds in scenarios where standard greedy decoding strategies often fail to synthesize the correct answer.

\paragraph{Robustness to Noise.}
The MultiHop-RAG dataset is designed to test resilience against noisy and irrelevant context. On this benchmark, our method demonstrates superior robustness. It achieves an EM score of 35.3. In comparison, the advanced selection baselines perform significantly worse, with RAG + MMR achieving 32.1 and Filco scoring 29.8. This performance gap validates the effectiveness of our MMKP context selector. By explicitly modeling redundancy and information density constraints, our selector effectively filters out red herring documents. These distractions typically degrade the performance of standard RAG models.

\paragraph{Comparison with Agentic Baselines.}
We observe competitive performance from baselines such as CRAG and Self-RAG on single-hop tasks like PopQA. For instance, CRAG achieves the top score on PopQA with an EM of 45.5. However, our method demonstrates superior consistency across diverse difficulty levels. On HotpotQA, our approach remains robust. We achieve an F1 score of 49.1, which is comparable to the 51.2 scored by Self-RAG. More importantly, our method maintains higher consistency in retrieval-dependent generation, as evidenced by the Retrieval metrics discussed in the following subsection.

\subsection{Retrieval Quality and Context Optimization}
Table~\ref{tab:retrieval_performance} reports the Recall@5 performance. Our MMKP-based context selector demonstrates a clear advantage over both traditional ranking methods, such as Naive and RRR, and greedy diversity methods like MMR.

\paragraph{Effectiveness of MMKP.}
Self-Correcting RAG achieves the highest average recall of 72.0\% across all datasets. When compared directly to the RAG+MMR baseline, which has an average recall of 63.3\%, our method yields an absolute improvement of 8.7\%. This empirical evidence highlights the theoretical superiority of our approach. We formulate context selection as a constrained knapsack problem rather than relying on a greedy iterative process. By jointly optimizing for relevance and diversity within a strict token budget, MMKP retains crucial evidence that greedy methods often discard prematurely.

\paragraph{Handling Information Scarcity.}
The \textbf{HotpotQA} dataset relies heavily on bridging entities across documents to answer questions correctly. On this benchmark, our method reaches a recall of 93.6\%. This score is significantly higher than standard baselines and exceeds the strongest competitor, CRAG, which scores 91.5\%. This result suggests that the redundancy penalty in our MMKP formulation successfully diversifies the context window. It ensures that complementary document pairs required for multi-hop inference are both selected and preserved.

\section{Discussion}
\label{sec:discussion}

\subsection{Ablation Study}
To validate the effectiveness of the Self-Correcting RAG framework, we conducted ablation experiments by isolating the MMKP Context Selector and the NLI-Guided MCTS Generator. Table~\ref{tab:ablation_single} presents the comparative results across QA performance, retrieval quality, and faithfulness metrics.

\paragraph{Impact of MMKP Context Selection.}
Replacing the standard Top-$k$ retrieval with our MMKP formulation significantly boosts retrieval performance. Specifically, Recall@5 improves dramatically from 49.6\% to 71.8\%. This substantial gain confirms that modeling context selection as a multidimensional knapsack problem is highly effective. It reduces redundancy, such as filtering duplicate passages, and maximizes information density within the constrained token budget of 1500 tokens. However, better context alone is insufficient for perfect generation. While MMKP improves the presence of correct answers, raising the EM score to 34.5, the faithfulness metrics remain comparable to the baseline with an Attribution Precision (AP) of 0.58. This suggests that improved retrieval does not inherently prevent the generator from hallucinating ungrounded claims.

\paragraph{Impact of NLI-Guided MCTS.}
The integration of the NLI-Guided MCTS generator drastically improves the faithfulness of the generated content. It increases the Attribution Precision (AP) to 0.85 and significantly reduces the Contradiction Rate (CR) to 0.04. These metrics demonstrate that the NLI reward signal effectively penalizes reasoning paths that contradict retrieved evidence. The full \textit{Self-Correcting RAG} framework combines the strengths of both components. By grounding the generation process in high-quality, diverse context, it achieves the highest overall performance with an EM of 37.1 and an F1 score of 45.8.

\begin{table}[t]
\centering
\resizebox{\columnwidth}{!}{%
\begin{tabular}{lcccccc}
\toprule
\multirow{2}{*}{\textbf{Method}} & \multicolumn{2}{c}{\textbf{QA}} & \textbf{Retriev.} & \multicolumn{3}{c}{\textbf{Faithfulness}} \\
\cmidrule(lr){2-3} \cmidrule(lr){4-4} \cmidrule(lr){5-7}
& EM & F1 & Recall@5 & AP $\uparrow$ & CR $\downarrow$ & Sup \\
\midrule
Standard RAG & 25.8 & 36.1 & 49.6 & 0.52 & 0.15 & 0.65 \\
\midrule
 \quad w/ MMKP Only & 34.5 & 42.3 & \textbf{71.8} & 0.58 & 0.13 & 0.71 \\
\quad w/ MCTS Only & 31.2 & 39.7 & 50.1 & 0.82 & 0.06 & 0.84 \\
\midrule
\textbf{Self-Correcting} & \textbf{37.1} & \textbf{45.8} & \textbf{72.0} & \textbf{0.85} & \textbf{0.04} & \textbf{0.88} \\
\bottomrule
\end{tabular}%
}

\caption{Ablations. We report the average performance across all datasets to evaluate the individual contribution of the MMKP selector and MCTS generator.}
\label{tab:ablation_single}
\end{table}

\subsection{Sensitivity Analysis}
We analyze the robustness of our model against key hyperparameters defined in our configuration. We specifically investigate the impact of redundancy constraints in the MMKP selector and the computational budget allocated to the MCTS planner. Detailed numerical results and visualizations for these sensitivity analyses are provided in Appendix~\ref{app:sensitivity}.

\paragraph{Token and Redundancy Budgets.}
The MMKP selector relies on a critical balance between relevance and diversity. Reducing the redundancy budget, denoted as $C_{red}$, forces the model to select semantically distinct chunks. We observe that setting the scaled $C_{red}$ to approximately 120 yields optimal recall. Stricter budgets tend to discard relevant but lexically similar evidence.

\paragraph{MCTS Simulation Depth.}

The performance of the generator is closely tied to the MCTS search parameters. Increasing the number of simulations, $N$, improves the consistency of the generated answers. Our experiments indicate that a branching factor of $k=3$ combined with a maximum depth of 3 yields optimal reasoning accuracy without overcomplicating the search space. Furthermore, the penalty for contradiction proves crucial; reducing the magnitude of this penalty leads to a measurably higher rate of hallucinated content.

\subsection{Qualitative Analysis}
To investigate how Self-Correcting RAG rectifies errors, we analyze specific failure cases of the baseline. A common failure mode in multi-hop QA is reasoning shortcuts triggered by context crowding. In these instances, redundant retrieved chunks displace key evidence, causing the model to hallucinate connections based on parametric memory. A detailed step-by-step trace of this behavior is provided in Appendix~\ref{app:qualitative}.

\section{Conclusion}
We presented Self-Correcting RAG, a unified framework to enhance RAG robustness. We reformulated context selection as the Multidimensional Multi-choice Knapsack Problem (MMKP) to maximize information density under strict token budgets. Additionally, we proposed an NLI-guided MCTS generator for faithfulness, using NLI as a reward model to prune hallucinatory reasoning paths. Experiments on six benchmarks show it outperforms strong agentic baselines, especially in complex multi-hop tasks. However, the planner’s iterative nature increases inference latency; future work will optimize sample efficiency to reduce test-time search computational cost.

\section*{Limitations}
\label{sec:limitations}

Our proposed framework demonstrates notable improvements in faithfulness and reasoning capability. However, it entails limitations inherent to Self-Correcting RAG architectures. The primary constraint is the increased computational overhead and inference latency. Our MCTS-based approach differs from standard single-pass RAG. It requires performing multiple forward passes and NLI verifications for each reasoning step. We implement pareto-pruning for the MMKP selector to maintain polynomial time complexity. Nevertheless, the test-time search increases the time-to-first-token. This characteristic limits the current iteration's applicability in ultra-low-latency real-time scenarios.

Furthermore, the robustness of our reward mechanism relies on the quality of the off-the-shelf NLI model (e.g., RoBERTa-large-mnli). These auxiliary models are generally effective for standard domains. However, they may overlook subtle contradictions in highly specialized fields, such as law or medicine. Addressing this requires domain-specific fine-tuning. Finally, our MMKP selector aims to maximize information density. It operates on the assumption that redundancy within retrieved groups is semantically uniform. In cases of high semantic complexity, rigorous filtering might inadvertently discard complementary minority opinions.

\section*{Ethical Considerations}

Our work aims to enhance the reliability of Large Language Models. We focus on reducing hallucinations through constrained context selection and logical verification. By grounding generation in traceable evidence, we mitigate risks associated with disseminating factually incorrect information. However, we acknowledge the environmental impact of the test-time compute paradigm. The iterative nature of Monte Carlo Tree Search increases GPU utilization compared to standard decoding methods. Consequently, this results in higher energy consumption. Future work should focus on optimizing the planner's sample efficiency to reduce this computational footprint.

Additionally, as a Retrieval-Augmented Generation system, our model's outputs are constrained by the retrieved corpora. The results reflect the quality and potential biases of the source documents. Our NLI guidance penalizes logical contradictions. However, it does not verify the intrinsic factual accuracy of the retrieved text. If the knowledge base contains biased or toxic content, the system may reproduce these issues. It is important to note that faithfulness in this context refers to adherence to the retrieved context. It does not necessarily equate to absolute objective truth.

\section*{Acknowledgements}
This work was supported by the National Natural Science Foundation of China (No. 52578347).


\bibliography{main}

\begin{thebibliography}{65}
\providecommand{\natexlab}[1]{#1}

\bibitem[{Ahn et~al.(2022)Ahn, Brohan, Brown, Chebotar, Cortes, David, Finn, Fu, Gopalakrishnan, Hausman, Herzog, Ho, Hsu, Ibarz, Ichter, Irpan, Jang, Ruano, Jeffrey, Jesmonth, Joshi, Julian, Kalashnikov, Kuang, Lee, Levine, Lu, Luu, Parada, Pastor, Quiambao, Rao, Rettinghouse, Reyes, Sermanet, Sievers, Tan, Toshev, Vanhoucke, Xia, Xiao, Xu, Xu, Yan, and Zeng}]{ahn2022can}
Michael Ahn, Anthony Brohan, Noah Brown, Yevgen Chebotar, Omar Cortes, Byron David, Chelsea Finn, Chuyuan Fu, Keerthana Gopalakrishnan, Karol Hausman, Alex Herzog, Daniel Ho, Jasmine Hsu, Julian Ibarz, Brian Ichter, Alex Irpan, Eric Jang, Rosario~Jauregui Ruano, Kyle Jeffrey, and 26 others. 2022.
\newblock \href {https://arxiv.org/abs/2204.01691} {Do as i can, not as i say: Grounding language in robotic affordances}.
\newblock \emph{Preprint}, arXiv:2204.01691.

\bibitem[{Asai et~al.(2023)Asai, Wu, Wang, Sil, and Hajishirzi}]{asai2024selfrag}
Akari Asai, Zeqiu Wu, Yizhong Wang, Avirup Sil, and Hannaneh Hajishirzi. 2023.
\newblock \href {https://arxiv.org/abs/2310.11511} {Self-rag: Learning to retrieve, generate, and critique through self-reflection}.
\newblock \emph{Preprint}, arXiv:2310.11511.

\bibitem[{Besta et~al.(2024)Besta, Blach, Kubicek, Gerstenberger, Podstawski, Gianinazzi, Gajda, Lehmann, Niewiadomski, Nyczyk, and Hoefler}]{besta2024graph}
Maciej Besta, Nils Blach, Ales Kubicek, Robert Gerstenberger, Michal Podstawski, Lukas Gianinazzi, Joanna Gajda, Tomasz Lehmann, Hubert Niewiadomski, Piotr Nyczyk, and Torsten Hoefler. 2024.
\newblock \href {https://doi.org/10.1609/aaai.v38i16.29720} {Graph of thoughts: Solving elaborate problems with large language models}.
\newblock \emph{Proceedings of the AAAI Conference on Artificial Intelligence}, 38(16):17682--17690.

\bibitem[{Borgeaud et~al.(2022)Borgeaud, Mensch, Hoffmann, Cai, Rutherford, Millican, Van Den~Driessche, Lespiau, Damoc, Clark, De~Las~Casas, Guy, Menick, Ring, Hennigan, Huang, Maggiore, Jones, Cassirer, Brock, Paganini, Irving, Vinyals, Osindero, Simonyan, Rae, Elsen, and Sifre}]{borgeaud2022improving}
Sebastian Borgeaud, Arthur Mensch, Jordan Hoffmann, Trevor Cai, Eliza Rutherford, Katie Millican, George~Bm Van Den~Driessche, Jean-Baptiste Lespiau, Bogdan Damoc, Aidan Clark, Diego De~Las~Casas, Aurelia Guy, Jacob Menick, Roman Ring, Tom Hennigan, Saffron Huang, Loren Maggiore, Chris Jones, Albin Cassirer, and 9 others. 2022.
\newblock \href {https://proceedings.mlr.press/v162/borgeaud22a.html} {Improving language models by retrieving from trillions of tokens}.
\newblock In \emph{Proceedings of the 39th International Conference on Machine Learning}, volume 162 of \emph{Proceedings of Machine Learning Research}, pages 2206--2240. PMLR.

\bibitem[{Bubeck et~al.(2023)Bubeck, Chandrasekaran, Eldan, Gehrke, Horvitz, Kamar, Lee, Lee, Li, Lundberg, Nori, Palangi, Ribeiro, and Zhang}]{bubeck2023sparks}
Sébastien Bubeck, Varun Chandrasekaran, Ronen Eldan, Johannes Gehrke, Eric Horvitz, Ece Kamar, Peter Lee, Yin~Tat Lee, Yuanzhi Li, Scott Lundberg, Harsha Nori, Hamid Palangi, Marco~Tulio Ribeiro, and Yi~Zhang. 2023.
\newblock \href {https://arxiv.org/abs/2303.12712} {Sparks of artificial general intelligence: Early experiments with gpt-4}.
\newblock \emph{Preprint}, arXiv:2303.12712.

\bibitem[{Gao et~al.(2023)Gao, Ma, Lin, and Callan}]{gao2023precise}
Luyu Gao, Xueguang Ma, Jimmy Lin, and Jamie Callan. 2023.
\newblock \href {https://doi.org/10.18653/v1/2023.acl-long.99} {Precise zero-shot dense retrieval without relevance labels}.
\newblock In \emph{Proceedings of the 61st Annual Meeting of the Association for Computational Linguistics (Volume 1: Long Papers)}, pages 1762--1777, Toronto, Canada. Association for Computational Linguistics.

\bibitem[{Guo et~al.(2025)Guo, Wang, Guo, Li, Song, Tan, Liu, Bian, and Yang}]{guo2023connecting}
Qingyan Guo, Rui Wang, Junliang Guo, Bei Li, Kaitao Song, Xu~Tan, Guoqing Liu, Jiang Bian, and Yujiu Yang. 2025.
\newblock \href {https://arxiv.org/abs/2309.08532} {Evoprompt: Connecting llms with evolutionary algorithms yields powerful prompt optimizers}.
\newblock \emph{Preprint}, arXiv:2309.08532.

\bibitem[{Guu et~al.(2020)Guu, Lee, Tung, Pasupat, and Chang}]{guu2020realm}
Kelvin Guu, Kenton Lee, Zora Tung, Panupong Pasupat, and Mingwei Chang. 2020.
\newblock \href {https://proceedings.mlr.press/v119/guu20a.html} {Retrieval augmented language model pre-training}.
\newblock In \emph{Proceedings of the 37th International Conference on Machine Learning}, volume 119 of \emph{Proceedings of Machine Learning Research}, pages 3929--3938. PMLR.

\bibitem[{Hao et~al.(2023)Hao, Gu, Ma, Hong, Wang, Wang, and Hu}]{hao2023reasoning}
Shibo Hao, Yi~Gu, Haodi Ma, Joshua Hong, Zhen Wang, Daisy Wang, and Zhiting Hu. 2023.
\newblock \href {https://doi.org/10.18653/v1/2023.emnlp-main.507} {Reasoning with language model is planning with world model}.
\newblock In \emph{Proceedings of the 2023 Conference on Empirical Methods in Natural Language Processing}, pages 8154--8173, Singapore. Association for Computational Linguistics.

\bibitem[{He et~al.(2024)He, Zhong, Cai, Lee, and He}]{wang2024rest}
Zhenyu He, Zexuan Zhong, Tianle Cai, Jason Lee, and Di~He. 2024.
\newblock \href {https://doi.org/10.18653/v1/2024.naacl-long.88} {{REST}: Retrieval-based speculative decoding}.
\newblock In \emph{Proceedings of the 2024 Conference of the North American Chapter of the Association for Computational Linguistics: Human Language Technologies (Volume 1: Long Papers)}, pages 1582--1595, Mexico City, Mexico. Association for Computational Linguistics.

\bibitem[{Ho et~al.(2020)Ho, Nguyen, Sugawara, and Aizawa}]{ho20202wikimultihopqa}
Xanh Ho, Anh-Khoa~Duong Nguyen, Saku Sugawara, and Akiko Aizawa. 2020.
\newblock \href {https://arxiv.org/abs/2011.01060} {Constructing a multi-hop qa dataset for comprehensive evaluation of reasoning steps}.
\newblock \emph{Preprint}, arXiv:2011.01060.

\bibitem[{Hu et~al.(2025)Hu, Zhang, Jiang, Zhang, Wei, and Qing}]{zhang2024drag}
Wentao Hu, Wengyu Zhang, Yiyang Jiang, Chen~Jason Zhang, Xiaoyong Wei, and Li~Qing. 2025.
\newblock \href {https://doi.org/10.18653/v1/2025.acl-long.770} {Removal of hallucination on hallucination: Debate-augmented {RAG}}.
\newblock In \emph{Proceedings of the 63rd Annual Meeting of the Association for Computational Linguistics (Volume 1: Long Papers)}, pages 15839--15853, Vienna, Austria. Association for Computational Linguistics.

\bibitem[{Huang et~al.(2025)Huang, Yu, Ma, Zhong, Feng, Wang, Chen, Peng, Feng, Qin, and Liu}]{huang2023survey}
Lei Huang, Weijiang Yu, Weitao Ma, Weihong Zhong, Zhangyin Feng, Haotian Wang, Qianglong Chen, Weihua Peng, Xiaocheng Feng, Bing Qin, and Ting Liu. 2025.
\newblock \href {https://doi.org/10.1145/3703155} {A survey on hallucination in large language models: Principles, taxonomy, challenges, and open questions}.
\newblock \emph{ACM Trans. Inf. Syst.}, 43(2).

\bibitem[{Izacard et~al.(2023)Izacard, Lewis, Lomeli, Hosseini, Petroni, Schick, Dwivedi-Yu, Joulin, Riedel, and Grave}]{izacard2023atlas}
Gautier Izacard, Patrick Lewis, Maria Lomeli, Lucas Hosseini, Fabio Petroni, Timo Schick, Jane Dwivedi-Yu, Armand Joulin, Sebastian Riedel, and Edouard Grave. 2023.
\newblock \href {http://jmlr.org/papers/v24/23-0037.html} {Atlas: Few-shot learning with retrieval augmented language models}.
\newblock \emph{Journal of Machine Learning Research}, 24(251):1--43.

\bibitem[{Jagerman et~al.(2023)Jagerman, Zhuang, Qin, Wang, and Bendersky}]{jagerman2023query}
Rolf Jagerman, Honglei Zhuang, Zhen Qin, Xuanhui Wang, and Michael Bendersky. 2023.
\newblock \href {https://arxiv.org/abs/2305.03653} {Query expansion by prompting large language models}.
\newblock \emph{Preprint}, arXiv:2305.03653.

\bibitem[{Ji et~al.(2023)Ji, Lee, Frieske, Yu, Su, Xu, Ishii, Bang, Madotto, and Fung}]{ji2023survey}
Ziwei Ji, Nayeon Lee, Rita Frieske, Tiezheng Yu, Dan Su, Yan Xu, Etsuko Ishii, Ye~Jin Bang, Andrea Madotto, and Pascale Fung. 2023.
\newblock \href {https://doi.org/10.1145/3571730} {Survey of hallucination in natural language generation}.
\newblock \emph{ACM Comput. Surv.}, 55(12).

\bibitem[{Jiang et~al.(2024)Jiang, Wu, Luo, Li, Lin, Yang, and Qiu}]{wang2023longllmlingua}
Huiqiang Jiang, Qianhui Wu, Xufang Luo, Dongsheng Li, Chin-Yew Lin, Yuqing Yang, and Lili Qiu. 2024.
\newblock \href {https://doi.org/10.18653/v1/2024.acl-long.91} {{L}ong{LLML}ingua: Accelerating and enhancing {LLM}s in long context scenarios via prompt compression}.
\newblock In \emph{Proceedings of the 62nd Annual Meeting of the Association for Computational Linguistics (Volume 1: Long Papers)}, pages 1658--1677, Bangkok, Thailand. Association for Computational Linguistics.

\bibitem[{Jiang et~al.(2023)Jiang, Xu, Gao, Sun, Liu, Dwivedi-Yu, Yang, Callan, and Neubig}]{jiang2023active}
Zhengbao Jiang, Frank Xu, Luyu Gao, Zhiqing Sun, Qian Liu, Jane Dwivedi-Yu, Yiming Yang, Jamie Callan, and Graham Neubig. 2023.
\newblock \href {https://doi.org/10.18653/v1/2023.emnlp-main.495} {Active retrieval augmented generation}.
\newblock In \emph{Proceedings of the 2023 Conference on Empirical Methods in Natural Language Processing}, pages 7969--7992, Singapore. Association for Computational Linguistics.

\bibitem[{Kojima et~al.(2022)Kojima, Gu, Reid, Matsuo, and Iwasawa}]{kojima2022large}
Takeshi Kojima, Shixiang~(Shane) Gu, Machel Reid, Yutaka Matsuo, and Yusuke Iwasawa. 2022.
\newblock \href {https://proceedings.neurips.cc/paper_files/paper/2022/file/8bb0d291acd4acf06ef112099c16f326-Paper-Conference.pdf} {Large language models are zero-shot reasoners}.
\newblock In \emph{Advances in Neural Information Processing Systems}, volume~35, pages 22199--22213. Curran Associates, Inc.

\bibitem[{Kwiatkowski et~al.(2019)Kwiatkowski, Palomaki, Redfield, Collins, Parikh, Alberti, Epstein, Polosukhin, Devlin, Lee, Toutanova, Jones, Kelcey, Chang, Dai, Uszkoreit, Le, and Petrov}]{kwiatkowski2019natural}
Tom Kwiatkowski, Jennimaria Palomaki, Olivia Redfield, Michael Collins, Ankur Parikh, Chris Alberti, Danielle Epstein, Illia Polosukhin, Jacob Devlin, Kenton Lee, Kristina Toutanova, Llion Jones, Matthew Kelcey, Ming-Wei Chang, Andrew~M. Dai, Jakob Uszkoreit, Quoc Le, and Slav Petrov. 2019.
\newblock \href {https://doi.org/10.1162/tacl_a_00276} {Natural questions: A benchmark for question answering research}.
\newblock \emph{Transactions of the Association for Computational Linguistics}, 7:453--466.

\bibitem[{Lewis et~al.(2020)Lewis, Perez, Piktus, Petroni, Karpukhin, Goyal, K\"{u}ttler, Lewis, Yih, Rockt\"{a}schel, Riedel, and Kiela}]{lewis2020retrieval}
Patrick Lewis, Ethan Perez, Aleksandra Piktus, Fabio Petroni, Vladimir Karpukhin, Naman Goyal, Heinrich K\"{u}ttler, Mike Lewis, Wen-tau Yih, Tim Rockt\"{a}schel, Sebastian Riedel, and Douwe Kiela. 2020.
\newblock \href {https://proceedings.neurips.cc/paper_files/paper/2020/file/6b493230205f780e1bc26945df7481e5-Paper.pdf} {Retrieval-augmented generation for knowledge-intensive nlp tasks}.
\newblock In \emph{Advances in Neural Information Processing Systems}, volume~33, pages 9459--9474. Curran Associates, Inc.

\bibitem[{Li et~al.(2023{\natexlab{a}})Li, Cheng, Zhao, Nie, and Wen}]{li2024halueval}
Junyi Li, Xiaoxue Cheng, Wayne~Xin Zhao, Jian-Yun Nie, and Ji-Rong Wen. 2023{\natexlab{a}}.
\newblock \href {https://arxiv.org/abs/2305.11747} {Halueval: A large-scale hallucination evaluation benchmark for large language models}.
\newblock \emph{Preprint}, arXiv:2305.11747.

\bibitem[{Li et~al.(2023{\natexlab{b}})Li, Zhao, Yu, Song, Li, Yu, Li, Huang, and Li}]{li2023api}
Minghao Li, Yingxiu Zhao, Bowen Yu, Feifan Song, Hangyu Li, Haiyang Yu, Zhoujun Li, Fei Huang, and Yongbin Li. 2023{\natexlab{b}}.
\newblock \href {https://arxiv.org/abs/2304.08244} {Api-bank: A comprehensive benchmark for tool-augmented llms}.
\newblock \emph{Preprint}, arXiv:2304.08244.

\bibitem[{Liu et~al.(2023)Liu, Jiang, Zhang, Liu, Zhang, Biswas, and Stone}]{liu2023llm}
Bo~Liu, Yuqian Jiang, Xiaohan Zhang, Qiang Liu, Shiqi Zhang, Joydeep Biswas, and Peter Stone. 2023.
\newblock \href {https://arxiv.org/abs/2304.11477} {Llm+p: Empowering large language models with optimal planning proficiency}.
\newblock \emph{Preprint}, arXiv:2304.11477.

\bibitem[{Long(2023)}]{long2023large}
Jieyi Long. 2023.
\newblock \href {https://arxiv.org/abs/2305.08291} {Large language model guided tree-of-thought}.
\newblock \emph{Preprint}, arXiv:2305.08291.

\bibitem[{Ma et~al.(2023{\natexlab{a}})Ma, Gong, He, Zhao, and Duan}]{ma2023query}
Xinbei Ma, Yeyun Gong, Pengcheng He, Hai Zhao, and Nan Duan. 2023{\natexlab{a}}.
\newblock \href {https://doi.org/10.18653/v1/2023.emnlp-main.322} {Query rewriting in retrieval-augmented large language models}.
\newblock In \emph{Proceedings of the 2023 Conference on Empirical Methods in Natural Language Processing}, pages 5303--5315, Singapore. Association for Computational Linguistics.

\bibitem[{Ma et~al.(2023{\natexlab{b}})Ma, Zhang, Pradeep, and Lin}]{ma2023zero}
Xueguang Ma, Xinyu Zhang, Ronak Pradeep, and Jimmy Lin. 2023{\natexlab{b}}.
\newblock \href {https://arxiv.org/abs/2305.02156} {Zero-shot listwise document reranking with a large language model}.
\newblock \emph{Preprint}, arXiv:2305.02156.

\bibitem[{Mallen et~al.(2023)Mallen, Asai, Zhong, Das, Khashabi, and Hajishirzi}]{mallen2023popqa}
Alex Mallen, Akari Asai, Victor Zhong, Rajarshi Das, Daniel Khashabi, and Hannaneh Hajishirzi. 2023.
\newblock \href {https://doi.org/10.18653/v1/2023.acl-long.546} {When not to trust language models: Investigating effectiveness of parametric and non-parametric memories}.
\newblock In \emph{Proceedings of the 61st Annual Meeting of the Association for Computational Linguistics (Volume 1: Long Papers)}, pages 9802--9822, Toronto, Canada. Association for Computational Linguistics.

\bibitem[{Manakul et~al.(2023)Manakul, Liusie, and Gales}]{manakul2023selfcheckgpt}
Potsawee Manakul, Adian Liusie, and Mark Gales. 2023.
\newblock \href {https://doi.org/10.18653/v1/2023.emnlp-main.557} {{S}elf{C}heck{GPT}: Zero-resource black-box hallucination detection for generative large language models}.
\newblock In \emph{Proceedings of the 2023 Conference on Empirical Methods in Natural Language Processing}, pages 9004--9017, Singapore. Association for Computational Linguistics.

\bibitem[{Mialon et~al.(2023)Mialon, Dessì, Lomeli, Nalmpantis, Pasunuru, Raileanu, Rozière, Schick, Dwivedi-Yu, Celikyilmaz, Grave, LeCun, and Scialom}]{mialon2023augmented}
Grégoire Mialon, Roberto Dessì, Maria Lomeli, Christoforos Nalmpantis, Ram Pasunuru, Roberta Raileanu, Baptiste Rozière, Timo Schick, Jane Dwivedi-Yu, Asli Celikyilmaz, Edouard Grave, Yann LeCun, and Thomas Scialom. 2023.
\newblock \href {https://arxiv.org/abs/2302.07842} {Augmented language models: a survey}.
\newblock \emph{Preprint}, arXiv:2302.07842.

\bibitem[{Min et~al.(2023)Min, Krishna, Lyu, Lewis, Yih, Koh, Iyyer, Zettlemoyer, and Hajishirzi}]{min2023factscore}
Sewon Min, Kalpesh Krishna, Xinxi Lyu, Mike Lewis, Wen-tau Yih, Pang Koh, Mohit Iyyer, Luke Zettlemoyer, and Hannaneh Hajishirzi. 2023.
\newblock \href {https://doi.org/10.18653/v1/2023.emnlp-main.741} {{FA}ct{S}core: Fine-grained atomic evaluation of factual precision in long form text generation}.
\newblock In \emph{Proceedings of the 2023 Conference on Empirical Methods in Natural Language Processing}, pages 12076--12100, Singapore. Association for Computational Linguistics.

\bibitem[{Muennighoff et~al.(2023)Muennighoff, Tazi, Magne, and Reimers}]{muennighoff2023mteb}
Niklas Muennighoff, Nouamane Tazi, Loic Magne, and Nils Reimers. 2023.
\newblock \href {https://doi.org/10.18653/v1/2023.eacl-main.148} {{MTEB}: Massive text embedding benchmark}.
\newblock In \emph{Proceedings of the 17th Conference of the European Chapter of the Association for Computational Linguistics}, pages 2014--2037, Dubrovnik, Croatia. Association for Computational Linguistics.

\bibitem[{OpenAI et~al.(2024)OpenAI, Achiam, Adler, Agarwal, Ahmad, Akkaya, Aleman, Almeida, Altenschmidt, Altman, Anadkat, Avila, Babuschkin, Balaji, Balcom, Baltescu, Bao, Bavarian, Belgum, Bello, Berdine, Bernadett-Shapiro, Berner, Bogdonoff, Boiko, Boyd, Brakman, Brockman, Brooks, Brundage, Button, Cai, Campbell, Cann, Carey, Carlson, Carmichael, Chan, Chang, Chantzis, Chen, Chen, Chen, Chen, Chen, Chess, Cho, Chu, Chung, Cummings, Currier, Dai, Decareaux, Degry, Deutsch, Deville, Dhar, Dohan, Dowling, Dunning, Ecoffet, Eleti, Eloundou, Farhi, Fedus, Felix, Fishman, Forte, Fulford, Gao, Georges, Gibson, Goel, Gogineni, Goh, Gontijo-Lopes, Gordon, Grafstein, Gray, Greene, Gross, Gu, Guo, Hallacy, Han, Harris, He, Heaton, Heidecke, Hesse, Hickey, Hickey, Hoeschele, Houghton, Hsu, Hu, Hu, Huizinga, Jain, Jain, Jang, Jiang, Jiang, Jin, Jin, Jomoto, Jonn, Jun, Kaftan, Łukasz Kaiser, Kamali, Kanitscheider, Keskar, Khan, Kilpatrick, Kim, Kim, Kim, Kirchner, Kiros, Knight, Kokotajlo, Łukasz Kondraciuk,
  Kondrich, Konstantinidis, Kosic, Krueger, Kuo, Lampe, Lan, Lee, Leike, Leung, Levy, Li, Lim, Lin, Lin, Litwin, Lopez, Lowe, Lue, Makanju, Malfacini, Manning, Markov, Markovski, Martin, Mayer, Mayne, McGrew, McKinney, McLeavey, McMillan, McNeil, Medina, Mehta, Menick, Metz, Mishchenko, Mishkin, Monaco, Morikawa, Mossing, Mu, Murati, Murk, Mély, Nair, Nakano, Nayak, Neelakantan, Ngo, Noh, Ouyang, O'Keefe, Pachocki, Paino, Palermo, Pantuliano, Parascandolo, Parish, Parparita, Passos, Pavlov, Peng, Perelman, de~Avila Belbute~Peres, Petrov, de~Oliveira~Pinto, Michael, Pokorny, Pokrass, Pong, Powell, Power, Power, Proehl, Puri, Radford, Rae, Ramesh, Raymond, Real, Rimbach, Ross, Rotsted, Roussez, Ryder, Saltarelli, Sanders, Santurkar, Sastry, Schmidt, Schnurr, Schulman, Selsam, Sheppard, Sherbakov, Shieh, Shoker, Shyam, Sidor, Sigler, Simens, Sitkin, Slama, Sohl, Sokolowsky, Song, Staudacher, Such, Summers, Sutskever, Tang, Tezak, Thompson, Tillet, Tootoonchian, Tseng, Tuggle, Turley, Tworek, Uribe, Vallone,
  Vijayvergiya, Voss, Wainwright, Wang, Wang, Wang, Ward, Wei, Weinmann, Welihinda, Welinder, Weng, Weng, Wiethoff, Willner, Winter, Wolrich, Wong, Workman, Wu, Wu, Wu, Xiao, Xu, Yoo, Yu, Yuan, Zaremba, Zellers, Zhang, Zhang, Zhao, Zheng, Zhuang, Zhuk, and Zoph}]{openai2023gpt4}
OpenAI, Josh Achiam, Steven Adler, Sandhini Agarwal, Lama Ahmad, Ilge Akkaya, Florencia~Leoni Aleman, Diogo Almeida, Janko Altenschmidt, Sam Altman, Shyamal Anadkat, Red Avila, Igor Babuschkin, Suchir Balaji, Valerie Balcom, Paul Baltescu, Haiming Bao, Mohammad Bavarian, Jeff Belgum, and 262 others. 2024.
\newblock \href {https://arxiv.org/abs/2303.08774} {Gpt-4 technical report}.
\newblock \emph{Preprint}, arXiv:2303.08774.

\bibitem[{Patil et~al.(2024)Patil, Zhang, Wang, and Gonzalez}]{patil2023gorilla}
Shishir~G. Patil, Tianjun Zhang, Xin Wang, and Joseph~E. Gonzalez. 2024.
\newblock \href {https://doi.org/10.52202/079017-4020} {Gorilla: Large language model connected with massive apis}.
\newblock In \emph{Advances in Neural Information Processing Systems}, volume~37, pages 126544--126565. Curran Associates, Inc.

\bibitem[{Pradeep et~al.(2023)Pradeep, Sharifymoghaddam, and Lin}]{pradeep2023rankvicuna}
Ronak Pradeep, Sahel Sharifymoghaddam, and Jimmy Lin. 2023.
\newblock \href {https://arxiv.org/abs/2309.15088} {Rankvicuna: Zero-shot listwise document reranking with open-source large language models}.
\newblock \emph{Preprint}, arXiv:2309.15088.

\bibitem[{Qin et~al.(2023)Qin, Liang, Ye, Zhu, Yan, Lu, Lin, Cong, Tang, Qian, Zhao, Hong, Tian, Xie, Zhou, Gerstein, Li, Liu, and Sun}]{qin2024toolllm}
Yujia Qin, Shihao Liang, Yining Ye, Kunlun Zhu, Lan Yan, Yaxi Lu, Yankai Lin, Xin Cong, Xiangru Tang, Bill Qian, Sihan Zhao, Lauren Hong, Runchu Tian, Ruobing Xie, Jie Zhou, Mark Gerstein, Dahai Li, Zhiyuan Liu, and Maosong Sun. 2023.
\newblock \href {https://arxiv.org/abs/2307.16789} {Toolllm: Facilitating large language models to master 16000+ real-world apis}.
\newblock \emph{Preprint}, arXiv:2307.16789.

\bibitem[{Ram et~al.(2023)Ram, Levine, Dalmedigos, Muhlgay, Shashua, Leyton-Brown, and Shoham}]{ram2023context}
Ori Ram, Yoav Levine, Itay Dalmedigos, Dor Muhlgay, Amnon Shashua, Kevin Leyton-Brown, and Yoav Shoham. 2023.
\newblock \href {https://doi.org/10.1162/tacl_a_00605} {In-context retrieval-augmented language models}.
\newblock \emph{Transactions of the Association for Computational Linguistics}, 11:1316--1331.

\bibitem[{Romera-Paredes et~al.(2024)Romera-Paredes, Barekatain, Novikov, Balog, Kumar, Dupont, Ruiz, Ellenberg, Wang, Fawzi et~al.}]{romeraparedes2024mathematical}
Bernardino Romera-Paredes, Mohammadamin Barekatain, Alexander Novikov, Matej Balog, M~Pawan Kumar, Emilien Dupont, Francisco~JR Ruiz, Jordan~S Ellenberg, Pengming Wang, Omar Fawzi, and 1 others. 2024.
\newblock \href {https://doi.org/10.1038/s41586-023-06924-6} {Mathematical discoveries from program search with large language models}.
\newblock \emph{Nature}, 625(7995):468--475.

\bibitem[{Ruan et~al.(2023)Ruan, Chen, Zhang, Xu, Bao, du~qing, shi shiwei, Mao, Zeng, and Zhao}]{ruan2023tptu}
Jingqing Ruan, YiHong Chen, Bin Zhang, Zhiwei Xu, Tianpeng Bao, du~qing, shi shiwei, Hangyu Mao, Xingyu Zeng, and Rui Zhao. 2023.
\newblock \href {https://openreview.net/forum?id=GrkgKtOjaH} {{TPTU}: Task planning and tool usage of large language model-based {AI} agents}.
\newblock In \emph{NeurIPS 2023 Foundation Models for Decision Making Workshop}.

\bibitem[{Sarthi et~al.(2024)Sarthi, Abdullah, Tuli, Khanna, Goldie, and Manning}]{sarthi2024raptor}
Parth Sarthi, Salman Abdullah, Aditi Tuli, Shubh Khanna, Anna Goldie, and Christopher~D Manning. 2024.
\newblock \href {https://openreview.net/forum?id=GN921JHCRw} {{RAPTOR}: Recursive abstractive processing for tree-organized retrieval}.
\newblock In \emph{The Twelfth International Conference on Learning Representations}.

\bibitem[{Schick et~al.(2023)Schick, Dwivedi-Yu, Dessi, Raileanu, Lomeli, Hambro, Zettlemoyer, Cancedda, and Scialom}]{schick2023toolformer}
Timo Schick, Jane Dwivedi-Yu, Roberto Dessi, Roberta Raileanu, Maria Lomeli, Eric Hambro, Luke Zettlemoyer, Nicola Cancedda, and Thomas Scialom. 2023.
\newblock \href {https://proceedings.neurips.cc/paper_files/paper/2023/file/d842425e4bf79ba039352da0f658a906-Paper-Conference.pdf} {Toolformer: Language models can teach themselves to use tools}.
\newblock In \emph{Advances in Neural Information Processing Systems}, volume~36, pages 68539--68551. Curran Associates, Inc.

\bibitem[{Shen et~al.(2023)Shen, Song, Tan, Li, Lu, and Zhuang}]{shen2024hugginggpt}
Yongliang Shen, Kaitao Song, Xu~Tan, Dongsheng Li, Weiming Lu, and Yueting Zhuang. 2023.
\newblock \href {https://proceedings.neurips.cc/paper_files/paper/2023/file/77c33e6a367922d003ff102ffb92b658-Paper-Conference.pdf} {Hugginggpt: Solving ai tasks with chatgpt and its friends in hugging face}.
\newblock In \emph{Advances in Neural Information Processing Systems}, volume~36, pages 38154--38180. Curran Associates, Inc.

\bibitem[{Shinn et~al.(2023)Shinn, Cassano, Gopinath, Narasimhan, and Yao}]{shinn2024reflexion}
Noah Shinn, Federico Cassano, Ashwin Gopinath, Karthik Narasimhan, and Shunyu Yao. 2023.
\newblock \href {https://proceedings.neurips.cc/paper_files/paper/2023/file/1b44b878bb782e6954cd888628510e90-Paper-Conference.pdf} {Reflexion: language agents with verbal reinforcement learning}.
\newblock In \emph{Advances in Neural Information Processing Systems}, volume~36, pages 8634--8652. Curran Associates, Inc.

\bibitem[{Sun et~al.(2024)Sun, Yan, Ma, Wang, Ren, Chen, Yin, and Ren}]{sun2023chatgpt}
Weiwei Sun, Lingyong Yan, Xinyu Ma, Shuaiqiang Wang, Pengjie Ren, Zhumin Chen, Dawei Yin, and Zhaochun Ren. 2024.
\newblock \href {https://arxiv.org/abs/2304.09542} {Is chatgpt good at search? investigating large language models as re-ranking agents}.
\newblock \emph{Preprint}, arXiv:2304.09542.

\bibitem[{Tang and Yang(2024)}]{tang2024multihoprag}
Yixuan Tang and Yi~Yang. 2024.
\newblock \href {https://arxiv.org/abs/2401.15391} {Multihop-rag: Benchmarking retrieval-augmented generation for multi-hop queries}.
\newblock \emph{Preprint}, arXiv:2401.15391.

\bibitem[{Touvron et~al.(2023)Touvron, Martin, Stone, Albert, Almahairi, Babaei, Bashlykov, Batra, Bhargava, Bhosale, Bikel, Blecher, Ferrer, Chen, Cucurull, Esiobu, Fernandes, Fu, Fu, Fuller, Gao, Goswami, Goyal, Hartshorn, Hosseini, Hou, Inan, Kardas, Kerkez, Khabsa, Kloumann, Korenev, Koura, Lachaux, Lavril, Lee, Liskovich, Lu, Mao, Martinet, Mihaylov, Mishra, Molybog, Nie, Poulton, Reizenstein, Rungta, Saladi, Schelten, Silva, Smith, Subramanian, Tan, Tang, Taylor, Williams, Kuan, Xu, Yan, Zarov, Zhang, Fan, Kambadur, Narang, Rodriguez, Stojnic, Edunov, and Scialom}]{touvron2023llama}
Hugo Touvron, Louis Martin, Kevin Stone, Peter Albert, Amjad Almahairi, Yasmine Babaei, Nikolay Bashlykov, Soumya Batra, Prajjwal Bhargava, Shruti Bhosale, Dan Bikel, Lukas Blecher, Cristian~Canton Ferrer, Moya Chen, Guillem Cucurull, David Esiobu, Jude Fernandes, Jeremy Fu, Wenyin Fu, and 49 others. 2023.
\newblock \href {https://arxiv.org/abs/2307.09288} {Llama 2: Open foundation and fine-tuned chat models}.
\newblock \emph{Preprint}, arXiv:2307.09288.

\bibitem[{Trivedi et~al.(2022)Trivedi, Balasubramanian, Khot, and Sabharwal}]{trivedi2022musique}
Harsh Trivedi, Niranjan Balasubramanian, Tushar Khot, and Ashish Sabharwal. 2022.
\newblock \href {https://doi.org/10.1162/tacl_a_00475} {Musique: Multihop questions via single-hop question composition}.
\newblock \emph{Transactions of the Association for Computational Linguistics}, 10:539--554.

\bibitem[{Trivedi et~al.(2023)Trivedi, Balasubramanian, Khot, and Sabharwal}]{liu2023cotrag}
Harsh Trivedi, Niranjan Balasubramanian, Tushar Khot, and Ashish Sabharwal. 2023.
\newblock \href {https://doi.org/10.18653/v1/2023.acl-long.557} {Interleaving retrieval with chain-of-thought reasoning for knowledge-intensive multi-step questions}.
\newblock In \emph{Proceedings of the 61st Annual Meeting of the Association for Computational Linguistics (Volume 1: Long Papers)}, pages 10014--10037, Toronto, Canada. Association for Computational Linguistics.

\bibitem[{Turpin et~al.(2023)Turpin, Michael, Perez, and Bowman}]{turpin2024language}
Miles Turpin, Julian Michael, Ethan Perez, and Samuel Bowman. 2023.
\newblock \href {https://proceedings.neurips.cc/paper_files/paper/2023/file/ed3fea9033a80fea1376299fa7863f4a-Paper-Conference.pdf} {Language models don\textquotesingle t always say what they think: Unfaithful explanations in chain-of-thought prompting}.
\newblock In \emph{Advances in Neural Information Processing Systems}, volume~36, pages 74952--74965. Curran Associates, Inc.

\bibitem[{Wang et~al.(2024)Wang, Yang, Huang, Jiao, Yang, Jiang, Majumder, and Wei}]{wang2022text}
Liang Wang, Nan Yang, Xiaolong Huang, Binxing Jiao, Linjun Yang, Daxin Jiang, Rangan Majumder, and Furu Wei. 2024.
\newblock \href {https://arxiv.org/abs/2212.03533} {Text embeddings by weakly-supervised contrastive pre-training}.
\newblock \emph{Preprint}, arXiv:2212.03533.

\bibitem[{Wang et~al.(2023{\natexlab{a}})Wang, Wei, Schuurmans, Le, Chi, Narang, Chowdhery, and Zhou}]{wang2023self}
Xuezhi Wang, Jason Wei, Dale Schuurmans, Quoc Le, Ed~Chi, Sharan Narang, Aakanksha Chowdhery, and Denny Zhou. 2023{\natexlab{a}}.
\newblock \href {https://arxiv.org/abs/2203.11171} {Self-consistency improves chain of thought reasoning in language models}.
\newblock \emph{Preprint}, arXiv:2203.11171.

\bibitem[{Wang et~al.(2023{\natexlab{b}})Wang, Araki, Jiang, Parvez, and Neubig}]{wu2024filco}
Zhiruo Wang, Jun Araki, Zhengbao Jiang, Md~Rizwan Parvez, and Graham Neubig. 2023{\natexlab{b}}.
\newblock \href {https://arxiv.org/abs/2311.08377} {Learning to filter context for retrieval-augmented generation}.
\newblock \emph{Preprint}, arXiv:2311.08377.

\bibitem[{Wei et~al.(2022)Wei, Wang, Schuurmans, Bosma, ichter, Xia, Chi, Le, and Zhou}]{wei2022chain}
Jason Wei, Xuezhi Wang, Dale Schuurmans, Maarten Bosma, brian ichter, Fei Xia, Ed~Chi, Quoc~V Le, and Denny Zhou. 2022.
\newblock \href {https://proceedings.neurips.cc/paper_files/paper/2022/file/9d5609613524ecf4f15af0f7b31abca4-Paper-Conference.pdf} {Chain-of-thought prompting elicits reasoning in large language models}.
\newblock In \emph{Advances in Neural Information Processing Systems}, volume~35, pages 24824--24837. Curran Associates, Inc.

\bibitem[{Xiao et~al.(2024)Xiao, Liu, Zhang, Muennighoff, Lian, and Nie}]{xiao2023cpack}
Shitao Xiao, Zheng Liu, Peitian Zhang, Niklas Muennighoff, Defu Lian, and Jian-Yun Nie. 2024.
\newblock \href {https://doi.org/10.1145/3626772.3657878} {C-pack: Packed resources for general chinese embeddings}.
\newblock In \emph{Proceedings of the 47th International ACM SIGIR Conference on Research and Development in Information Retrieval}, SIGIR '24, page 641–649, New York, NY, USA. Association for Computing Machinery.

\bibitem[{Xu et~al.(2023)Xu, Shi, and Choi}]{xu2024recomp}
Fangyuan Xu, Weijia Shi, and Eunsol Choi. 2023.
\newblock \href {https://arxiv.org/abs/2310.04408} {Recomp: Improving retrieval-augmented lms with compression and selective augmentation}.
\newblock \emph{Preprint}, arXiv:2310.04408.

\bibitem[{Yan et~al.(2024)Yan, Gu, Zhu, and Ling}]{yan2024corrective}
Shi-Qi Yan, Jia-Chen Gu, Yun Zhu, and Zhen-Hua Ling. 2024.
\newblock \href {https://openreview.net/forum?id=JnWJbrnaUE} {Corrective retrieval augmented generation}.

\bibitem[{Yang et~al.(2024)Yang, Wang, Lu, Liu, Le, Zhou, and Chen}]{yang2024large}
Chengrun Yang, Xuezhi Wang, Yifeng Lu, Hanxiao Liu, Quoc~V Le, Denny Zhou, and Xinyun Chen. 2024.
\newblock \href {https://openreview.net/forum?id=Bb4VGOWELI} {Large language models as optimizers}.
\newblock In \emph{The Twelfth International Conference on Learning Representations}.

\bibitem[{Yang et~al.(2018)Yang, Qi, Zhang, Bengio, Cohen, Salakhutdinov, and Manning}]{yang2018hotpotqa}
Zhilin Yang, Peng Qi, Saizheng Zhang, Yoshua Bengio, William Cohen, Ruslan Salakhutdinov, and Christopher~D. Manning. 2018.
\newblock \href {https://doi.org/10.18653/v1/D18-1259} {{H}otpot{QA}: A dataset for diverse, explainable multi-hop question answering}.
\newblock In \emph{Proceedings of the 2018 Conference on Empirical Methods in Natural Language Processing}, pages 2369--2380, Brussels, Belgium. Association for Computational Linguistics.

\bibitem[{Yao et~al.(2023{\natexlab{a}})Yao, Yu, Zhao, Shafran, Griffiths, Cao, and Narasimhan}]{yao2024tree}
Shunyu Yao, Dian Yu, Jeffrey Zhao, Izhak Shafran, Tom Griffiths, Yuan Cao, and Karthik Narasimhan. 2023{\natexlab{a}}.
\newblock \href {https://proceedings.neurips.cc/paper_files/paper/2023/file/271db9922b8d1f4dd7aaef84ed5ac703-Paper-Conference.pdf} {Tree of thoughts: Deliberate problem solving with large language models}.
\newblock In \emph{Advances in Neural Information Processing Systems}, volume~36, pages 11809--11822. Curran Associates, Inc.

\bibitem[{Yao et~al.(2023{\natexlab{b}})Yao, Zhao, Yu, Du, Shafran, Narasimhan, and Cao}]{yao2023react}
Shunyu Yao, Jeffrey Zhao, Dian Yu, Nan Du, Izhak Shafran, Karthik~R Narasimhan, and Yuan Cao. 2023{\natexlab{b}}.
\newblock \href {https://openreview.net/forum?id=WE_vluYUL-X} {React: Synergizing reasoning and acting in language models}.
\newblock In \emph{The Eleventh International Conference on Learning Representations}.

\bibitem[{Ye et~al.(2024)Ye, Wang, Cao, Berto, Hua, Kim, Park, and Song}]{ye2024reevo}
Haoran Ye, Jiarui Wang, Zhiguang Cao, Federico Berto, Chuanbo Hua, Haeyeon Kim, Jinkyoo Park, and Guojie Song. 2024.
\newblock \href {https://doi.org/10.52202/079017-1381} {Reevo: Large language models as hyper-heuristics with reflective evolution}.
\newblock In \emph{Advances in Neural Information Processing Systems}, volume~37, pages 43571--43608. Curran Associates, Inc.

\bibitem[{Zelikman et~al.(2022)Zelikman, Wu, Mu, and Goodman}]{zelikman2022star}
Eric Zelikman, Yuhuai Wu, Jesse Mu, and Noah Goodman. 2022.
\newblock \href {https://proceedings.neurips.cc/paper_files/paper/2022/file/639a9a172c044fbb64175b5fad42e9a5-Paper-Conference.pdf} {Star: Bootstrapping reasoning with reasoning}.
\newblock In \emph{Advances in Neural Information Processing Systems}, volume~35, pages 15476--15488. Curran Associates, Inc.

\bibitem[{Zhang et~al.(2025)Zhang, Li, Cui, Cai, Liu, Fu, Huang, Zhao, Zhang, Chen, Wang, Luu, Bi, Shi, and Shi}]{zhang2023siren}
Yue Zhang, Yafu Li, Leyang Cui, Deng Cai, Lemao Liu, Tingchen Fu, Xinting Huang, Enbo Zhao, Yu~Zhang, Yulong Chen, Longyue Wang, Anh~Tuan Luu, Wei Bi, Freda Shi, and Shuming Shi. 2025.
\newblock \href {https://doi.org/10.1162/COLI.a.16} {Siren’s song in the ai ocean: A survey on hallucination in large language models}.
\newblock \emph{Computational Linguistics}, pages 1--46.

\bibitem[{Zhou et~al.(2024)Zhou, Yan, Shlapentokh-Rothman, Wang, and Wang}]{zhou2024language}
Andy Zhou, Kai Yan, Michal Shlapentokh-Rothman, Haohan Wang, and Yu-Xiong Wang. 2024.
\newblock \href {https://arxiv.org/abs/2310.04406} {Language agent tree search unifies reasoning acting and planning in language models}.
\newblock \emph{Preprint}, arXiv:2310.04406.

\bibitem[{Zhou et~al.(2023)Zhou, Schärli, Hou, Wei, Scales, Wang, Schuurmans, Cui, Bousquet, Le, and Chi}]{zhou2023least}
Denny Zhou, Nathanael Schärli, Le~Hou, Jason Wei, Nathan Scales, Xuezhi Wang, Dale Schuurmans, Claire Cui, Olivier Bousquet, Quoc Le, and Ed~Chi. 2023.
\newblock \href {https://arxiv.org/abs/2205.10625} {Least-to-most prompting enables complex reasoning in large language models}.
\newblock \emph{Preprint}, arXiv:2205.10625.

\end{thebibliography}



\appendix

\section*{Appendices}

Within this supplementary material, we elaborate on the following aspects:

\begin{itemize}[leftmargin=*, label=$\bullet$]
    \item Appendix~\ref{app:datasets}: Dataset Details
    \item Appendix~\ref{sec:appendix_implementation}: Implementation Details and Hyperparameters
    \item Appendix~\ref{sec:appendix_main}: Theoretical Proofs and Analysis
    \item Appendix~\ref{app:Results}: Detailed Experimental Results
    \item Appendix~\ref{app:sensitivity}: Sensitivity Analysis Results
    \item Appendix~\ref{app:prompts}: LLM Prompts
    \item Appendix~\ref{app:qualitative}: Qualitative Analysis Case Study

\end{itemize}

\section{Dataset Details}
\label{app:datasets}
We evaluate our approach on six representative datasets covering Simple QA, Multi-Hop QA, and Multi-Document QA scenarios. The statistics of the evaluation datasets are summarized in Table~\ref{tab:dataset_stats}.

\paragraph{Simple QA.}
We first consider single-hop tasks that require retrieving facts from open-domain sources.
\begin{itemize}
    \item \textbf{Natural Questions (NQ):} An open-domain QA dataset. In our experiments, we use a subset of 1,000 queries and retrieve from a corpus of 9,633 passages.
    \item \textbf{PopQA:} This dataset focuses on long-tail entities which often require precise knowledge retrieval. We evaluate on 1,000 queries with a retrieval pool of 8,676 passages.
\end{itemize}

\paragraph{Multi-Hop QA.}
These datasets require the model to perform reasoning across multiple documents (typically 2--4 hops) to derive the answer.
\begin{itemize}
    \item \textbf{MuSiQue:} A challenging multi-hop dataset. Our evaluation involves 1,000 queries and the largest passage pool in our setup (11,656 passages), requiring complex reasoning chains.
    \item \textbf{2WikiMultiHopQA (2Wiki):} Based on Wikipedia, requiring 2--4 hops. We use 1,000 queries and 6,119 passages.
    \item \textbf{HotpotQA:} We utilize the distractor setting to test the model's ability to filter irrelevant information. The setup includes 1,000 queries and 9,811 passages.
\end{itemize}

\paragraph{Multi-Doc QA.}
Finally, we evaluate robustness against noise in a retrieval-augmented generation setting.
\begin{itemize}
    \item \textbf{MultiHop-RAG:} Designed to test noise robustness. Unlike the standard QA sets, this evaluation uses a larger set of 2,556 queries over 609 specific passages to measure the F1 score and retrieval accuracy.
\end{itemize}

\begin{table*}[t]
\centering
\small
\begin{tabular}{lcccccc}
\toprule
\textbf{Category} & \textbf{Dataset} & \textbf{Type} & \textbf{Avg. Hops} & \textbf{\# Queries} & \textbf{\# Passages} & \textbf{Metric} \\
\midrule
\multirow{2}{*}{\textbf{Simple QA}} 
 & NQ & Open-domain & 1 & 1,000 & 9,633 & EM / F1 \\
 & PopQA & Long-tail Entity & 1 & 1,000 & 8,676 & EM / F1 \\
\midrule
\multirow{3}{*}{\textbf{Multi-Hop QA}} 
 & MuSiQue & Multi-hop & 2.40 & 1,000 & 11,656 & EM / F1 \\
 & 2WikiMultiHopQA & Multi-hop & 2.42 & 1,000 & 6,119 & EM / F1 \\
 & HotpotQA & Multi-hop & 2 & 1,000 & 9,811 & EM / F1 \\
\midrule
\textbf{Multi-Doc QA} 
 & MultiHop-RAG & Noise Robustness & 2.70 & 2,556 & 609 & EM / F1 \\
\bottomrule
\end{tabular}
\caption{Statistics of the evaluation datasets. The "Queries" and "Passages" columns denote the specific counts used in our experimental evaluation.}
\label{tab:dataset_stats}
\end{table*}

\section{Implementation Details and Hyperparameters}
\label{sec:appendix_implementation}

\subsection{General Implementation Details}
We implemented our framework using PyTorch and the Hugging Face Transformers library. All experiments were conducted on a cluster of NVIDIA A100 (80GB) GPUs.
For the retrieval backbone, we utilized a hybrid approach combining dense and sparse retrieval. We employed \texttt{BAAI/bge-small-en-v1.5} as the dense embedding model and \texttt{scikit-learn}'s \texttt{TfidfVectorizer} (ngram range 1-2, max features 200k) for sparse retrieval. The results were fused using Reciprocal Rank Fusion (RRF) with $k=60$, alongside a centroid-based query expansion strategy where the query embedding is refined using the mean of the top-5 dense retrieval results.
For the generator, we utilized \texttt{Qwen2.5-7B-Instruct} as the backbone Large Language Model (LLM), loaded in \texttt{bfloat16} precision. The NLI verification signal was derived from \texttt{roberta-large-mnli}.

\subsection{Baseline Implementation Details}
\label{app:baselines}
To ensure a rigorous comparison, we aligned the base models and resource constraints across all baselines where applicable:

\begin{itemize}
    \item \textbf{Standard RAG (Naive):} We follow the standard "Retrieve-then-Generate" paradigm. We retrieve the top-$k$ ($k=5$) chunks using the same hybrid retriever (BGE + TF-IDF) described above and feed them directly into the \texttt{Qwen2.5-7B-Instruct} model. No advanced selection or re-ranking is applied.
    \item \textbf{RAG + MMR:} We implemented Maximal Marginal Relevance (MMR) to re-rank the retrieval candidates. We set the diversity hyperparameter $\lambda=0.6$, iteratively selecting documents that maximize a linear combination of query relevance and novelty relative to the already selected set, until the token budget (1500 tokens) is exhausted.
    \item \textbf{Self-RAG:} We utilized the official \texttt{selfrag/selfrag-llama2-7b} checkpoint. To maintain fairness in information access, we restricted the retrieved context size provided to Self-RAG to be approximately equivalent to our MMKP token budget ($\sim 5$ chunks or 1500 tokens).
\end{itemize}

\subsection{Our Method: Configuration and Logic}

\subsubsection{MMKP Context Selector}
The MMKP selector transforms the retrieval list into a constrained optimization problem. We first grouped retrieval candidates using a greedy clustering algorithm based on cosine similarity thresholds ($\tau_{sim} = 0.82$). For each candidate document $d_{ij}$ in group $G_i$, we calculated:
\begin{itemize}
    \item \textbf{Value ($v_{ij}$):} A weighted sum of relevance and diversity: $v_{ij} = \alpha \cdot \text{Score}_{\text{fusion}}(q, d_{ij}) + \beta \cdot (1 - \text{Sim}(d_{ij}, \mu_{G_i}))$, where $\mu_{G_i}$ is the group centroid.
    \item \textbf{Costs ($w_{ij}$):} A 2-dimensional cost vector containing the token length ($L_{token}$) and a semantic redundancy penalty ($C_{red}$). The redundancy penalty is calculated as the mean similarity to other group members, scaled by a factor of 100.
\end{itemize}
We solved this NP-hard problem using a Dynamic Programming approach with Pareto pruning to remove dominated states (states with higher costs and lower value), ensuring tractability.

\subsubsection{NLI-Guided MCTS Generator}
We implemented the generator as a Monte Carlo Tree Search (MCTS) process that explores the reasoning space at test time.
\begin{itemize}
    \item \textbf{Node Expansion:} At each step, the model chooses between two action types: \textit{Answer} (generate a response hypothesis) or \textit{Augment} (retrieve additional evidence using the current query).
    \item \textbf{Reward Function:} We employed \texttt{roberta-large-mnli} to compute a faithfulness reward. The reward $R$ is defined as $R = w_e \cdot P(entail) + w_n \cdot P(neutral) + w_c \cdot P(contradict)$. We set a severe penalty for contradictions ($w_c = -2.0$) to aggressively prune hallucinatory paths.
    \item \textbf{Search Strategy:} We used the UCT (Upper Confidence Bound for Trees) algorithm with an exploration constant $c_{ucb}=1.4$ to balance exploration of new reasoning paths and exploitation of high-reward trajectories.
\end{itemize}

\subsection{Hyperparameters}
The specific hyperparameters for the MMKP selector and MCTS generator used in our main experiments are detailed in Table~\ref{tab:mmkp_params} and Table~\ref{tab:mcts_params}.

\begin{table}[H]
\centering
\small
\begin{tabular}{lr}
\toprule
\textbf{Parameter} & \textbf{Value} \\
\midrule
Token Budget ($C_{\text{token}}$) & 1500 \\
Redundancy Budget ($C_{\text{red}}$) & 120 \\
Relevance Weight ($\alpha$) & 0.7 \\
Diversity Weight ($\beta$) & 0.3 \\
Similarity Threshold ($\tau_{\text{sim}}$) & 0.82 \\
Redundancy Cost Scale & 100.0 \\
\bottomrule
\end{tabular}
\caption{Hyperparameters for the MMKP Context Selector. Weights were tuned on the HotpotQA validation set.}
\label{tab:mmkp_params}
\end{table}

\begin{table}[H]
\centering
\small
\begin{tabular}{lr}
\toprule
\textbf{Parameter} & \textbf{Value} \\
\midrule
\multicolumn{2}{c}{\textit{Generator (LLM)}} \\
Model & Qwen2.5-7B-Instruct \\
Temperature & 0.7 \\
Top-p & 0.9 \\
Max New Tokens & 256 \\
\midrule
\multicolumn{2}{c}{\textit{MCTS Search}} \\
Simulations ($N$) & 24 \\
Branching Factor ($k$) & 3 \\
Max Search Depth & 3 \\
Exploration Constant ($c_{ucb}$) & 1.4 \\
\midrule
\multicolumn{2}{c}{\textit{Reward Function (NLI)}} \\
Entailment Weight ($w_e$) & 1.0 \\
Neutral Weight ($w_n$) & -0.2 \\
Contradiction Weight ($w_c$) & -2.0 \\
\bottomrule
\end{tabular}
\caption{Hyperparameters for NLI-Guided MCTS. The generator uses these settings during the test-time search phase.}
\label{tab:mcts_params}
\end{table}

\section{Theoretical Proofs and Analysis}
\label{sec:appendix_main}

\subsection{Supplementary Definitions for MMKP}
\label{app:mmkp_defs}

\subsubsection{Multidimensional Cost Vectors}
Unlike the standard Knapsack problem (single weight), RAG systems face multiple resource constraints. We define the cost vector $\mathbf{w}_{ij} \in \mathbb{R}^K$ for each item. We model $K=2$ dimensions:
\begin{enumerate}
    \setlength{\itemsep}{0pt}  
    \item Token Consumption ($k=1$): $w_{ij}^{(1)} = \text{Len}(d_{ij})$.
    \item Redundancy Penalty ($k=2$): Derived from the intra-group centroid.
    \begin{equation}
    \label{eq:red_cost}
        \begin{split}
        w_{ij}^{(2)} &= \lambda_{\text{red}} \cdot \Bigg( \frac{1}{|G_i| - 1} \\
        &\quad \sum_{d \in G_i \setminus \{d_{ij}\}} \cos\bigl(\Phi(d_{ij}), \Phi(d)\bigr) \Bigg)
        \end{split}
    \end{equation}
\end{enumerate}
This cost dimension penalizes items that are too central or generic within their cluster, preferring unique information, scaled by $\lambda_{red}$.

\subsubsection{Utility Function}
The utility $v_{ij}$ balances query relevance and global diversity:
\begin{equation}
    \label{eq:mmkp_utility}
    v_{ij} = \alpha \cdot \mathcal{F}_{fusion}(q, d_{ij}) + \beta \cdot (1 - \text{Sim}(d_{ij}, \mu_{G_i}))
\end{equation}
where $\mathcal{F}_{fusion}$ incorporates both dense and sparse (TF-IDF) retrieval scores (via Reciprocal Rank Fusion), and $\mu_{G_i}$ is the centroid of group $G_i$.

\subsection{Computational Complexity of MMKP }
\label{app:proof_np_hard}

\textbf{Theorem 1 (NP-hardness).}\label{thm:np_hard}
\textit{The RAG Document Selection problem formulated as MMKP (Eq.~\ref{eq:mmkp_objective}) is NP-hard.}

\begin{proof}
We perform a reduction from the classic 0/1 Knapsack Problem, which is known to be NP-hard.
Let a standard Knapsack instance be defined by $n$ items with values $v_i$, weights $w_i$, and capacity $W$.
We construct a special instance of our RAG-MMKP as follows:
\begin{enumerate}
    \setlength{\itemsep}{0pt}  
    \item Create $m=n$ groups.
    \item Each group $G_i$ contains exactly two items: $\{d_{i,1}, d_{i,0}\}$.
    \item Item $d_{i,1}$ (representing ``taking'' item $i$) has value $v_i$ and weight vector $\mathbf{w}_{i,1} = [w_i, 0, \dots, 0]$.
    \item Item $d_{i,0}$ (representing ``leaving'' item $i$) has value $0$ and weight vector $\mathbf{0}$.
    \item Set the MMKP capacity vector $\mathbf{C} = [W, \infty, \dots, \infty]$.
\end{enumerate}
The constraint $\sum_{j} x_{ij} \le 1$ in MMKP forces the selection of exactly one item per group (either the real item or the dummy zero-cost item), which is mathematically equivalent to the binary choice $x_i \in \{0,1\}$ in the standard Knapsack problem.
Since 0/1 Knapsack is NP-hard, and it is a special case of RAG-MMKP, RAG-MMKP is NP-hard.
\end{proof}
\subsection{FPTAS for Core MMKP Variant}
\label{app:fptas_proof}

We analyze the single-dimensional variant ($D=1$) where each group has size 1 (equivalent to standard Knapsack). We prove the existence of a Fully Polynomial-Time Approximation Scheme (FPTAS).

\textbf{Theorem 2 (FPTAS).}\label{thm:fptas}
\textit{For any $\epsilon > 0$, there exists an algorithm that returns a solution $S$ such that
$V(S) \ge (1-\epsilon)\,OPT$ and runs in time polynomial in $n$ and $1/\epsilon$.}

\textbf{Algorithm Construction:}
\begin{enumerate}
    \setlength{\itemsep}{0pt}  
    \item Let $P = \max_i v_i$. Given error tolerance $\epsilon > 0$, define scaling factor $K = \frac{\epsilon P}{n}$.
    \item Define scaled values $v'_i = \lfloor \frac{v_i}{K} \rfloor$.
    \item Solve the problem using DP on values $v'_i$. The recurrence is:
    $DP[k, v] = \min(DP[k-1, v], w_k + DP[k-1, v - v'_k])$.
    The max possible scaled value is $V'_{max} \approx n \cdot \frac{P}{K} = \frac{n^2}{\epsilon}$.
\end{enumerate}

The algorithm is detailed in Algorithm~\ref{FPTAS}.

\begin{algorithm}[htb]
\caption{FPTAS for 0/1 Knapsack Problem}
\label{FPTAS}
\begin{algorithmic}[1] 
    \Procedure{FPTAS-Knapsack}{$v, w, C, \epsilon$}
        \State Let $n$ be the number of items.
        \State $P \gets \max_{i=1}^n v_i$
        \State $K \gets \frac{\epsilon P}{n}$
        \State For $i = 1, \dots, n: v'_i \gets \lfloor v_i/K \rfloor$
        \State $V'_{max} \gets \sum_{i=1}^n v'_i$
        \State Initialize DP table $T$ of size $V'_{max} + 1$ with $T[0] = 0$ and $T[p] = \infty$ for $p > 0$.
        \For{$i = 1, \dots, n$}
            \For{$p = V'_{max}, \dots, v'_i$}
                \State $T[p] \gets \min(T[p], w_i + T[p - v'_i])$
            \EndFor
        \EndFor
        \State Find the largest $p^*$ such that $T[p^*] \le C$.
        \State Reconstruct the set of items corresponding to $T[p^*]$ and return it.
    \EndProcedure
\end{algorithmic}
\end{algorithm}

\begin{proof}
Let $S^*$ be the optimbual set and $S$ be the set returned by our algorithm; we prove the approximation guarantee.
By definition of floor: $K v'_i \le v_i < K(v'_i + 1)$.
The optimal value $OPT = \sum_{i \in S^*} v_i$.
Our algorithm finds $S$ that optimizes the scaled values, so $\sum_{i \in S} v'_i \ge \sum_{i \in S^*} v'_i$.

Considering the lower bound of our solution $V(S)$, we have:
\begin{equation}
\begin{split}
    V(S) = \sum_{i \in S} v_i &\ge \sum_{i \in S} K v'_i = K \sum_{i \in S} v'_i \\
    &\ge K \sum_{i \in S^*} v'_i \\
    &> K \sum_{i \in S^*} \left(\frac{v_i}{K} - 1\right) \\
    &= \sum_{i \in S^*} v_i - \sum_{i \in S^*} K \\
    &= OPT - nK
\end{split}
\end{equation}

Substituting $K = \frac{\epsilon P}{n}$, we get $V(S) > OPT - \epsilon P$.
Since $OPT \ge P$, we have $V(S) \ge OPT - \epsilon OPT = (1-\epsilon)OPT$.

\textbf{Time Complexity:}
The DP table size is $O(n \cdot V'_{max}) = O\!\left(n \cdot \frac{n^2}{\epsilon}\right) = O\!\left(\frac{n^3}{\epsilon}\right)$.
This is polynomial in $n$ and $1/\epsilon$, satisfying the definition of FPTAS.
\end{proof}


\subsection{Heuristic DP Details and Pareto Pruning}
\label{app:pareto_pruning_impl}

\textbf{Pareto-Pruned Dynamic Programming.}
For the practical implementation where $D=2$, we utilize a Dynamic Programming approach with state pruning.
Let $DP[i]$ be the set of achievable states after considering group $G_i$. Each state is a tuple $(\mathbf{c}, v)$, representing the accumulated cost vector and value.
To avoid state explosion, we prune dominated states. A state $A=(\mathbf{c}_A, v_A)$ dominates $B=(\mathbf{c}_B, v_B)$ if and only if:
\begin{equation}
    \label{eq:pareto_dominate}
    v_A \ge v_B \quad \text{and} \quad \forall k: c_A^{(k)} \le c_B^{(k)}
\end{equation}
At each step $i$, we compute
\begin{equation}
    \label{eq:pareto_frontier}
    \begin{split}
        DP[i] =  \text{PF} & \Biggl(\bigl\{ (\mathbf{c} + \mathbf{w}_{ij}, v + v_{ij}) \mid (\mathbf{c}, v) \\
               & \quad  \in DP[i-1],\ d_{ij} \in G_i \bigr\}\Biggr)
    \end{split}
\end{equation}
This reduces the average complexity to polynomial time in practice.The algorithm is detailed in Algorithm~\ref{alg:pareto_dp}.

\begin{algorithm}[htb]
\caption{MMKP with Pareto-Pruned DP}
\label{alg:pareto_dp}
\begin{algorithmic}[1]
\State \textbf{Input:} Groups $\mathcal{G}$, Budgets $\mathbf{C}$
\State $DP \gets \{(0,0): 0.0\}$ \Comment{Map (cost\_tokens, cost\_red) $\to$ value}
\For{each group $G_i \in \mathcal{G}$}
    \State $DP_{new} \gets DP.\text{copy}()$
    \For{each item $d_{ij} \in G_i$}
        \For{each state $(\mathbf{c}, v) \in DP$}
            \State $\mathbf{c}' \gets \mathbf{c} + \mathbf{w}_{ij}$
            \State $v' \gets v + v_{ij}$
            \If{$\mathbf{c}' \preceq \mathbf{C}$}
                \State Update $DP_{new}$ with $(\mathbf{c}', v')$
            \EndIf
        \EndFor
    \EndFor
    \State $DP \gets \text{PruneDominated}(DP_{new})$
\EndFor
\State \Return $\max_{v} DP$
\end{algorithmic}
\end{algorithm}

The function \texttt{PruneDominated} removes any state $(\mathbf{c}_B, v_B)$ if there exists $(\mathbf{c}_A, v_A)$ such that $\mathbf{c}_A \le \mathbf{c}_B$ AND $v_A \ge v_B$. This ensures we only track the Pareto frontier.

\subsection{Detailed MCTS Search Procedure}
\label{app:mcts_details}

We employ a variant of the Upper Confidence Bound for Trees (UCT) \& PUCT-style selection. The search proceeds in four steps for $N_{sim}$ simulations:

\begin{enumerate}
    \setlength{\itemsep}{0pt}  
\item \textbf{Selection:} Starting from root $s_0$, we recursively select child nodes satisfying:
\begin{equation}
    \label{eq:puct_select}
    \begin{split}
        a^* = &\operatorname*{argmax}_{a \in \mathcal{A}(s)}  \Biggl( Q(s, a) \\
        & + c_{puct} \cdot P(a|s) \frac{\sqrt{N(s)}}{1 + N(s, a)} \Biggr)
    \end{split}
\end{equation}
where $Q(s,a)$ is the estimated value, $N(s)$ is the visit count, and $P(a|s)$ is the prior from the policy (LLM).

\item \textbf{Expansion:} If node $s$ is not a terminal state, we expand it by sampling $k$ candidate answers (Generative Actions) and optionally $m$ retrieval queries (Augmentative Actions).

\item \textbf{Simulation (Rollout):} From the expanded node, we perform a rollout using the base policy $\pi_\theta$ to generate a complete answer $y_{final}$.

\item \textbf{Backpropagation:} We compute the reward $R(y_{final})$ using Eq.~\ref{eq:nli_reward} and update the Q-values along the trajectory:
\begin{equation}
    \label{eq:q_update}
    Q(s, a) \leftarrow \frac{N(s, a) \cdot Q(s, a) + R}{N(s, a) + 1}
\end{equation}
\end{enumerate}

\subsection{Convergence Analysis of NLI-Guided MCTS}
\label{app:mcts_convergence}

We provide a theoretical justification for the use of UCT in the space of logical consistency.

\textbf{Theorem 3 (MCTS Consistency).}\label{thm:mcts_consistency}
\textit{As the number of simulations $N \to \infty$, the probability of selecting the optimal answer $y^*$ (defined as the answer maximizing the NLI reward) approaches 1.}

\begin{proof}

The RAG generation process is modeled as a finite-horizon MDP. The NLI reward $R(y) \in [-2, 1]$ is bounded.
The UCT selection rule is:
\begin{equation}
    \label{eq:uct_rule}
    \bar{X}_j + 2C_p \sqrt{\frac{2 \ln n}{n_j}}
\end{equation}
According to the Kocsis and Szepesvári (2006) theorem, UCT is consistent in finite-horizon domains. The regret $R_n$ after $n$ steps grows as $O(\ln n)$.
Specifically for our NLI-guided generation:
\begin{enumerate}
    \setlength{\itemsep}{0pt} 
    \item \textbf{Exploration:} The $w_{con} = -2.0$ penalty in our reward function (Eq.~\ref{eq:nli_reward}) acts as a soft pruning mechanism. Branches containing contradictions yield low Q-values.
    \item \textbf{Exploitation:} UCT exponentially allocates samples to branches with high entailment scores ($w_{ent}=1.0$).
\end{enumerate}
Therefore, provided the NLI model $\Theta$ is an approximate oracle of truth, the search policy $\pi_{MCTS}$ converges to the sentence sequence $y$ that maximizes logical entailment with the evidence set $\mathcal{E}$.
\end{proof}


\section{Detailed Experimental Results}
\label{app:Results}

We further investigate the source of our performance gains by analyzing the impact of token constraints and reasoning complexity.

\begin{figure}[htb]  
    \centering
    \includegraphics[width=0.9\linewidth]{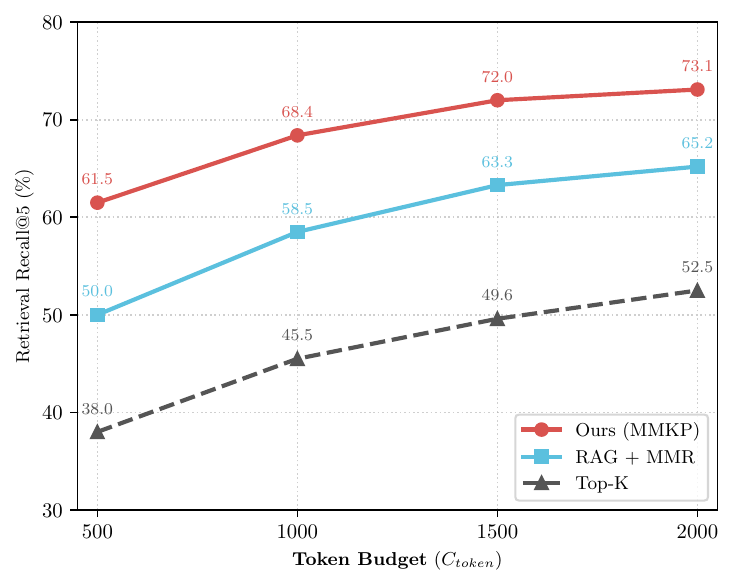}  
    
    \caption{Impact of token budget on retrieval performance. This figure compares the Retrieval Recall@5 of our MMKP method against RAG+MMR and Top-K baselines across varying token budgets ($C_{token}$). Our MMKP (red line) consistently outperforms the baselines, demonstrating superior robustness.}
    \label{fig:robust_hops1}
\end{figure}

\textbf{Token Budget Robustness.} Figure~\ref{fig:robust_hops1} demonstrates the retrieval recall across varying token constraints. While traditional Top-K selection suffers a severe performance drop to 38.0\% when the budget is restricted to 500 tokens, our MMKP selector exhibits superior resilience. It maintains a high recall of 61.5\% in this strict setting, outperforming Top-K by a margin of 23.5\% and the RAG+MMR baseline by 11.5\%. This substantial gap validates MMKP's ability to maximize information density when context space is scarce.

\begin{figure}[htb]  
    \centering
    \includegraphics[width=0.9\linewidth]{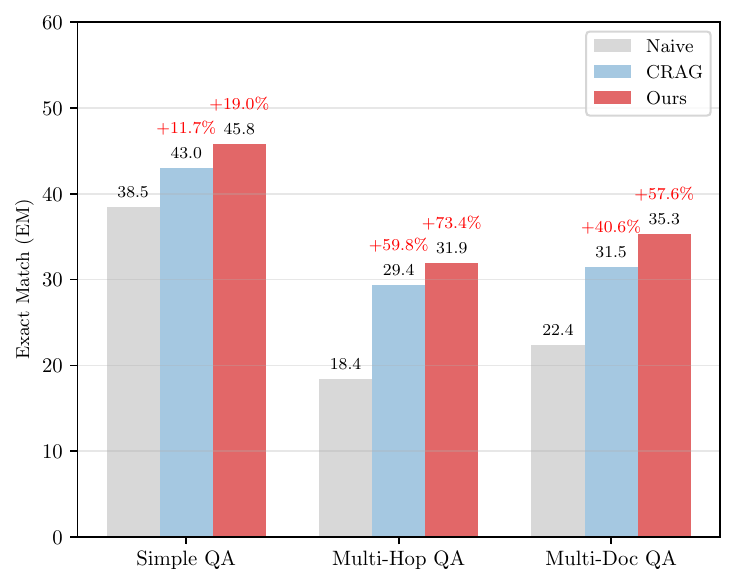}  
    
    \caption{Complexity Analysis. Our method shows significant scaling advantages on harder queries.}
    \label{fig:robust_hops2}
\end{figure}

\textbf{Reasoning on Hard Queries.} Figure~\ref{fig:robust_hops2} presents a performance decomposition based on query complexity. While our method attains a 19.0\% improvement over the Naive baseline on Simple QA, the performance gains are substantially amplified in complex reasoning scenarios. Specifically, on Multi-Hop QA, our approach demonstrates superior scaling by achieving a 73.4\% relative gain over the baseline and exceeding the CRAG model by 2.5 absolute points. Similarly, for Multi-Doc QA, our model records a 57.6\% improvement over the baseline. This result significantly widens the margin compared to CRAG, thereby validating the efficacy of our self-correction mechanism in handling long-context and multi-step reasoning tasks.

\section{Sensitivity Analysis Results}
\label{app:sensitivity}

In this appendix, we provide the detailed experimental results supporting the sensitivity analysis discussed in Section~\ref{sec:discussion}. We evaluate the impact of the redundancy budget on retrieval recall and the effect of MCTS simulation parameters on generation performance.

\subsection{Redundancy Budget in MMKP}
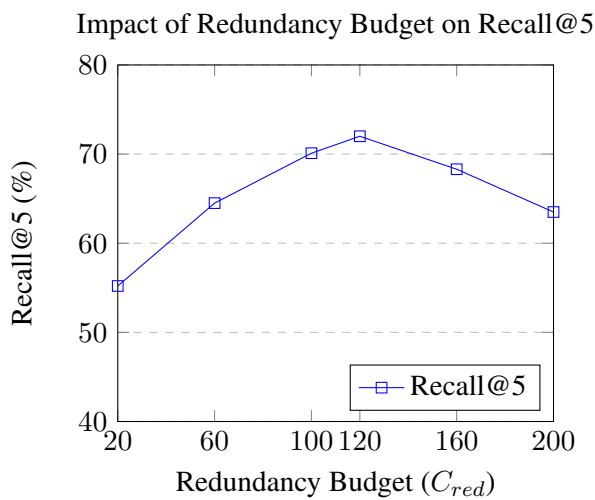
\begin{figure}[htb]
\centering
\begin{tikzpicture}
\begin{axis}[
    title={Impact of Redundancy Budget on Recall@5},
    xlabel={Redundancy Budget ($C_{red}$)},
    ylabel={Recall@5 (\%)},
    xmin=20, xmax=200,
    ymin=40, ymax=80,
    xtick={20, 60, 100, 120, 160, 200},
    ytick={40, 50, 60, 70, 80},
    legend pos=south east,
    ymajorgrids=true,
    grid style=dashed,
    width=0.95\columnwidth
]
\addplot[
    color=blue,
    mark=square,
    ]
    coordinates {
    (20, 55.2)(60, 64.5)(100, 70.1)(120, 72.0)(160, 68.3)(200, 63.5)
    };
    \legend{Recall@5}
\end{axis}
\end{tikzpicture}
\caption{Sensitivity of Recall@5 to the redundancy budget ($C_{red}$). A budget of approx. 120 provides the optimal trade-off, allowing sufficient diversity without excluding relevant evidence.}
\label{fig:redundancy_analysis}
\end{figure}

Figure~\ref{fig:redundancy_analysis} illustrates the relationship between the redundancy budget ($C_{red}$) and retrieval performance. The redundancy score is calculated based on maximum semantic similarity between selected passages.

\subsection{MCTS Simulation Parameters}
Table~\ref{tab:mcts_depth} details the generation performance (F1 Score) across different configurations of the MCTS planner. We vary the number of simulations ($N$) and the maximum search depth ($D$).

\begin{table}[h]
\centering
\resizebox{\columnwidth}{!}{%
\begin{tabular}{cccc}
\toprule
\textbf{Simulations ($N$)} & \textbf{Depth ($D$)} & \textbf{F1 Score} & \textbf{Faithfulness (AP)} \\
\midrule
1 (Greedy) & 1 & 36.1 & 0.52 \\
8 & 2 & 40.5 & 0.68 \\
16 & 3 & 43.2 & 0.79 \\
\rowcolor{lightblue}
\textbf{24} & \textbf{3} & \textbf{45.8} & \textbf{0.85} \\
32 & 4 & 46.0 & 0.86 \\
64 & 5 & 46.1 & 0.87 \\
\bottomrule
\end{tabular}%
}
\caption{Effect of MCTS hyperparameters on performance. $N$ denotes the number of simulations, and $D$ denotes the search depth. The chosen configuration ($N=24, D=3$) is highlighted.}
\label{tab:mcts_depth}
\end{table}

As observed, increasing $N$ beyond 24 yields diminishing returns in F1 score. Similarly, extending the depth beyond 3 provides marginal gains in faithfulness but complicates the reasoning trace unnecessarily for most tasks.

\section{LLM Prompts}
\label{app:prompts}

\begin{figure*}[t]
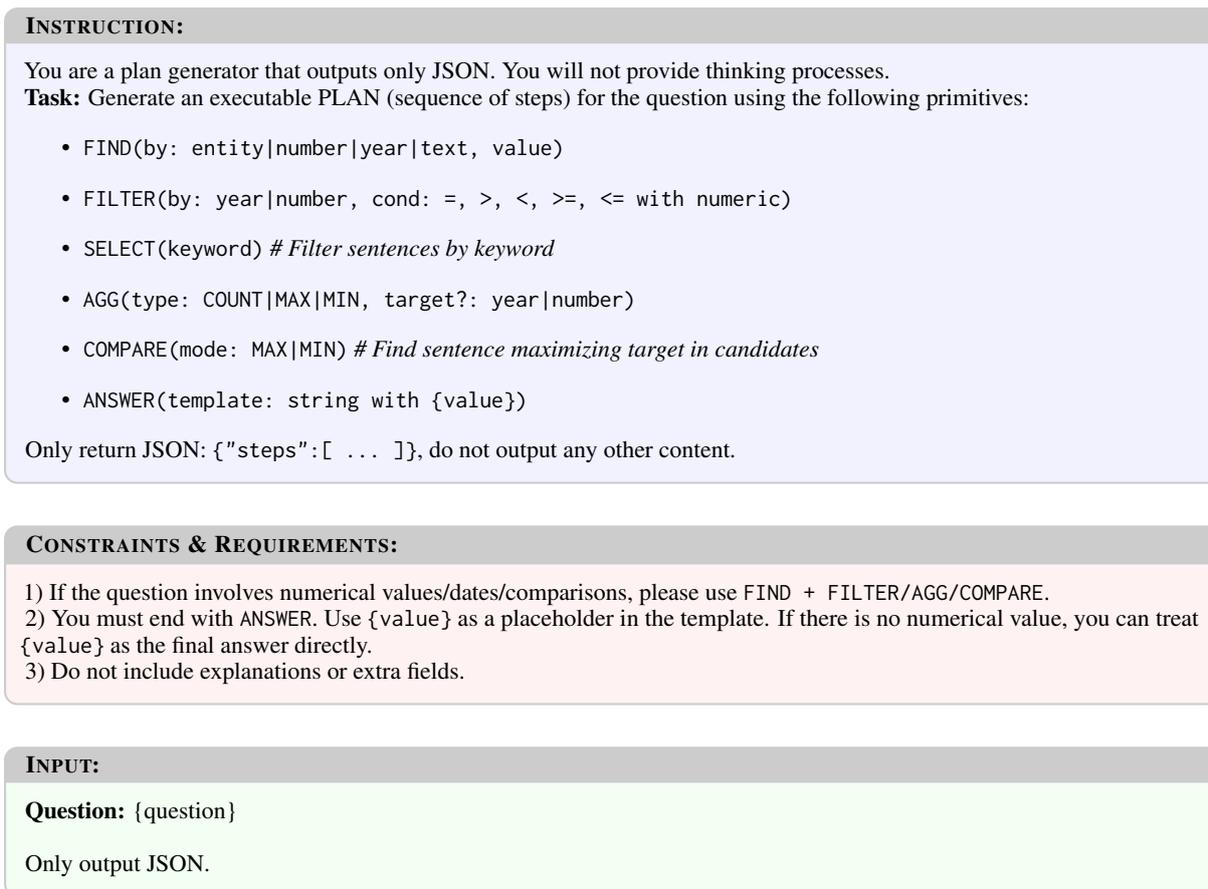

    \centering
    \small

    \begin{tcolorbox}[colback=blue!5, colframe=gray!40, title={\textsc{Instruction:}}, coltitle=black]
    You are a plan generator that outputs only JSON. You will not provide thinking processes. \\
    \textbf{Task:} Generate an executable PLAN (sequence of steps) for the question using the following primitives:
    \begin{itemize}
        \item \texttt{FIND(by: entity|number|year|text, value)}
        \item \texttt{FILTER(by: year|number, cond: =, >, <, >=, <= with numeric)}
        \item \texttt{SELECT(keyword)} \textit{\# Filter sentences by keyword}
        \item \texttt{AGG(type: COUNT|MAX|MIN, target?: year|number)}
        \item \texttt{COMPARE(mode: MAX|MIN)} \textit{\# Find sentence maximizing target in candidates}
        \item \texttt{ANSWER(template: string with \{value\})}
    \end{itemize}
    Only return JSON: \texttt{\{"steps":[ ... ]\}}, do not output any other content.
    \end{tcolorbox}
    
    \vspace{2mm}

    \begin{tcolorbox}[colback=red!5, colframe=gray!40, title={\textsc{Constraints \& Requirements:}}, coltitle=black]
    1) If the question involves numerical values/dates/comparisons, please use \texttt{FIND + FILTER/AGG/COMPARE}. \\
    2) You must end with \texttt{ANSWER}. Use \texttt{\{value\}} as a placeholder in the template. If there is no numerical value, you can treat \texttt{\{value\}} as the final answer directly. \\
    3) Do not include explanations or extra fields.
    \end{tcolorbox}

    \vspace{2mm}
    
    \begin{tcolorbox}[colback=green!5, colframe=gray!40, title={\textsc{Input:}}, coltitle=black]
    \textbf{Question:} \{question\} \\
    \\
    Only output JSON.
    \end{tcolorbox}

\caption{The prompt template used for the Symbolic Plan Generator. This prompt restricts the LLM to output a structured execution plan using predefined API primitives.}
\label{fig:planner_prompt}
\end{figure*}

We employ two distinct sets of prompts for our framework: one for the core Retrieval-Augmented Generation (RAG) tasks and another for the symbolic planning module. Figure~\ref{fig:rag_prompt} illustrates the prompts used for the NLI-Guided Generator, and Figure~\ref{fig:planner_prompt} shows the prompts for the Symbolic Plan Generator.

\begin{figure*}[htb]
    \centering
    \small

    \begin{tcolorbox}[colback=blue!5, colframe=gray!40, title={\textsc{Instruction:}}, coltitle=black]
    You are a rigorous multi-hop QA assistant. You must \textbf{answer strictly based on the provided document snippets}. Do not introduce external knowledge, subjective guesses, or information not mentioned in the documents. Please think deeply and answer according to the following steps:
    \begin{enumerate}
        \item \textbf{Decompose the Question:} Break down the core query points and the logical link of multi-hop reasoning (e.g., "Precondition $\to$ Intermediate Inference $\to$ Final Conclusion").
        \item \textbf{Locate Relevant Documents:} Examine candidate document snippets one by one, mark content directly/indirectly related to the core query, and record the corresponding \texttt{doc\_id}. Exclude irrelevant documents.
        \item \textbf{Integrate Information:} If relevant information is fragmented, logically connect it based on the document context. If there is no direct answer, provide a \textbf{speculative conclusion consistent with the document context} (do not return "unknown").
        \item \textbf{Verify Rationality:} Confirm that the answer originates entirely from the documents, contains no external information, and does not contradict the document content.
        \item \textbf{Output Result:} Provide a concise answer, the corresponding evidence \texttt{doc\_id}s, and explain the deep thinking process.
    \end{enumerate}
    \end{tcolorbox}
    
    \vspace{2mm}
    
    \begin{tcolorbox}[colback=green!5, colframe=gray!40, title={\textsc{Input Template:}}, coltitle=black]

    \textbf{Question:} \{question\} \\
    \\
    \textbf{Candidate Document Snippets} (Listed by \texttt{doc\_id}, potentially from different articles; content truncated): \\
    \{context\} \\
    \\
    Please strictly output in the following JSON format. The "reasoning" field must detail the deep thinking process (including question decomposition, document location, information integration, and rationality verification, at least 3 sentences):
    \begin{verbatim}
{
  "answer": "<Concise answer based on docs / Speculative conclusion>",
  "evidence_doc_ids": ["<doc_id1>", "<doc_id2>", "..."],
  "reasoning": "<Detailed thinking process...>"
}
    \end{verbatim}
    \end{tcolorbox}

\caption{The prompt template used for the NLI-Guided Generator. It enforces strict adherence to retrieved context and requires explicit reasoning steps to support the MCTS verification process.}
\label{fig:rag_prompt}
\end{figure*}

\section{Qualitative Analysis Case Studies}
\label{app:qualitative}

To investigate the mechanisms by which Self-Correcting RAG rectifies errors, we analyze specific failure cases of the baseline compared to our framework. We present two distinct scenarios: one focusing on context truncation due to redundancy, and another on attribute comparison amidst high-information noise.

\subsection{Case Study 1: Temporal Comparison with Context Truncation}
\label{sec:case_study_1}

In this first scenario, we examine a temporal comparison query regarding the founding dates of two magazines. The baseline fails due to context truncation of the second entity, leading to a hallucinated date. In contrast, our method rectifies this via (1) MMKP-based context de-duplication and (2) MCTS-guided verification.

Figure~\ref{fig:comprehensive_trace} illustrates a detailed distinct failure mode analysis. In the \textbf{Baseline (Left)}, the dense retriever retrieves three documents with high cosine similarity ($>0.88$) to the entity ``The American Conservative''. However, due to the limited context window (Top-3 constraints), these semantically redundant chunks crowd out the essential document containing the founding date of the second entity, ``The Weekly Standard''. Consequently, the CoT reasoner, lacking specific evidence, falls back on parametric memory, hallucinating an incorrect date (1950s) based on a broad ``Cold War era'' bias.

In contrast, our \textbf{Self-Correcting Framework (Right)} intervenes at two stages:
\begin{enumerate}
    \item \textbf{Context Construction:} The MMKP module clusters retrieved chunks by semantic intent. It identifies that Doc 1 and Doc 2 convey identical information (Cluster A) and discards the lower-ranked duplicate, effectively reserving token budget for the diverse Cluster B (Doc 3).
    \item \textbf{Reasoning Verification:} The MCTS planner expands the search space. When the model initially generates a hallucinated date (Branch $\pi_1$), the NLI verifier detects a low entailment score ($0.12$) against the retrieved context, assigning a negative reward ($r=-1.0$). This prompts the planner to backtrack and explore Branch $\pi_2$, which successfully extracts the correct date verified by high entailment ($0.98$), leading to the correct temporal comparison.
\end{enumerate}

\subsection{Case Study 2: Attribute Comparison under Information Noise}
\label{sec:case_study_2}

We further analyze a multi-hop reasoning scenario involving corporate history, which typically requires bridging entity relationships and precise attribute comparison. The query asks: \textit{``Who had a longer tenure as CEO of the company that acquired DeepMind: Eric Schmidt or Larry Page?''}

This query presents a specific challenge: ``DeepMind'' is a keyword heavily associated with recent AI breakthroughs, creating a ``distractor dense'' retrieval environment.

As shown in Figure~\ref{fig:deepmind_trace}, the \textbf{Baseline} suffers from \textit{topic drift}. The dense retriever prioritizes documents about DeepMind's technical achievements (AlphaGo, AlphaZero) due to high semantic similarity with the query entity. These redundant documents occupy the limited context window, crowding out the essential biographical data regarding the CEOs of the parent company (Google). Consequently, the model relies on a ``Founder Heuristic''---assuming the founder (Larry Page) served the longest---resulting in a hallucinated conclusion.

In contrast, our \textbf{Self-Correcting RAG} demonstrates robustness:
\begin{enumerate}
    \item \textbf{MMKP Filtering:} The Context Selector groups the ``AI Achievement'' documents into a single semantic cluster. It recognizes that selecting multiple documents from this cluster offers diminishing returns. It effectively prunes the redundancy (similar to the mechanism described in Section~\ref{sec:case_study_1}), freeing up token budget to retrieve the specific tenure dates for both Schmidt and Page.
    \item \textbf{MCTS Verification:} The generator initially attempts a heuristic guess (Page $>$ Schmidt). However, the NLI-guided verifier checks this against the retrieved evidence (Schmidt: 10 years vs. Page: 4 years) and penalizes the contradiction ($Q = -1.0$). The planner then backtracks to perform the correct arithmetic comparison, producing a faithful result.
\end{enumerate}

\begin{figure*}[t]
    \centering
    \small
    
    \begin{tcolorbox}[colback=bg_query, colframe=gray!40, title={\textsc{Input Query \& Complexity Analysis}}, coltitle=black]
        \textbf{Query:} \textit{"Which magazine was founded first, 'The American Conservative' (TAC) or 'The Weekly Standard' (TWS)?"}
        \par\noindent\rule{\textwidth}{0.4pt}
        \textbf{Reasoning Requirement:} 
        1. \textbf{Multi-hop Retrieval:} Must find $Date(TAC)$ AND $Date(TWS)$. 
        2. \textbf{Temporal Comparison:} Compute $\min(Date_{TAC}, Date_{TWS})$.
        \hfill \textit{Hardness: High (Ambiguity in entity dates)}
    \end{tcolorbox}

    \vspace{2mm}

    \begin{minipage}[t]{0.49\textwidth}
        \begin{tcolorbox}[
            colback=bg_fail, colframe=border_fail, 
            title={\textcolor{black}{\iconSearch\ Baseline: Standard RAG (Failure Trace)}}, 
            coltitle=border_fail,
            equal height group=ragtrace 
        ]
            
            \textbf{\textsc{Step 1: Dense Retrieval (Top-K)}} \\
            \textit{Retriever matches query embedding to chunks. Limit: $K=3$.}
            
            \begin{itemize}
                \setlength\itemsep{0em}
                \item \textbf{[Rank 1] Doc A:} "TAC is a bi-monthly journal... founded in \textbf{2002} by Pat Buchanan..." \\
                \footnotesize \textcolor{gray}{$\to$ Sim: 0.89 (High relevance to 'TAC')} \normalsize
                
                \item \textbf{[Rank 2] Doc B:} "TAC (Website) offers opinion pieces... launched in 2002..." \\
                \footnotesize \textcolor{gray}{$\to$ Sim: 0.88 (Redundant Semantic Cluster)} \normalsize
                
                \item \textbf{[Rank 3] Doc C:} "Pat Buchanan co-founded TAC to promote anti-war views..." \\
                \footnotesize \textcolor{gray}{$\to$ Sim: 0.86 (Redundant Semantic Cluster)} \normalsize
            \end{itemize}
            
            \vspace{1mm}
            \textbf{\textcolor{red}{CRITICAL ERROR:}} \textit{Context window full. The document containing "The Weekly Standard (1995)" (Rank 4, Sim 0.84) is \textbf{truncated}.}

            \par\noindent\rule{\textwidth}{0.4pt}

            \textbf{\textsc{Step 2: CoT Reasoning (Hallucination)}} \\
            \textit{LLM Input: Doc A, Doc B, Doc C.}
            
            \begin{itemize}
                \item \textbf{Thought 1:} "Founding date of TAC is 2002 (from Doc A)."
                \item \textbf{Thought 2:} "Searching for TWS date... Not in context."
                \item \textbf{Thought 3 (Parametric Memory):} "TWS is a famous neo-con magazine. It feels older, associated with the 90s or Cold War. Let's assume \textbf{1950s}."
                \item \textbf{Comparsion:} $1950s < 2002$.
            \end{itemize}

            \vspace{1mm}
            \begin{tcolorbox}[colback=white, colframe=border_fail, sharp corners]
                \centering
                \textbf{Final Prediction:} \\
                \textcolor{border_fail}{\textbf{The Weekly Standard (approx. 1950s)}} \\
                \ding{55} \textit{Factually Incorrect Reason}
            \end{tcolorbox}
        \end{tcolorbox}
    \end{minipage}
    \hfill
    \begin{minipage}[t]{0.49\textwidth}
        \begin{tcolorbox}[
            colback=bg_success, colframe=border_success, 
            title={\textcolor{black}{\iconFilter\ Ours: Self-Correcting RAG (Success Trace)}}, 
            coltitle=border_success,
            equal height group=ragtrace 
        ]
            
            \textbf{\textsc{Phase 1: MMKP Context Selector}} \\
            \textit{Goal: Maximize Diversity / Minimize Redundancy.}
            
            \begin{itemize}
                \setlength\itemsep{0em}
                \item \textbf{Cluster 1 (Topic: TAC):} \{Doc A, Doc B, Doc C\}
                \item \textbf{Cluster 2 (Topic: TWS):} \{Doc D (Rank 4)\}
                \item \textbf{Action:} Select representative from Cluster 1 (Doc A) $\to$ \textbf{Discard Doc B \& C}. $\to$ \textbf{Retrieve Doc D}.
            \end{itemize}
            \textbf{Optimized Context:} \{Doc A (2002), Doc D ("TWS... founded Sep 17, 1995")\}
             \vspace{0.5cm} 
            \par\noindent\rule{\textwidth}{0.4pt}

            \textbf{\textsc{Phase 2: MCTS Planner Reasoning}} \\
            \textit{Policy: $\pi(a|s)$, Reward: NLI(Context, Hypothesis).}
            
            \textbf{Node 0: Root State (Question)}
            
            \vspace{0.6cm}
            \textbf{$\hookrightarrow$ Branch 1: Greedy Generation (Hallucination)} \\
            \hspace*{2em} \textit{Hypothesis:} "TWS founded in 1955." \\
            \hspace*{2em} $\to$ \textbf{Verification (NLI):} Context (Doc D: 1995) vs Hypothesis (1955). \\
            \hspace*{2em} $\to$ \textbf{Result:} \textbf{Contradiction} (0.99). \\
            \hspace*{2em} $\to$ \textbf{Reward $Q(s,a)$:} \textcolor{red}{\textbf{-1.0}} \textit{(Prune path)}

            \vspace{0.6cm}
            \textbf{$\hookrightarrow$ Branch 2: Guided Extraction (Correct)} \\
            \hspace*{2em} \textit{Hypothesis:} "TWS founded in 1995." \\
            \hspace*{2em} $\to$ \textbf{Verification (NLI):} Context (Doc D: 1995) vs Hypothesis (1995). \\
            \hspace*{2em} $\to$ \textbf{Result:} \textbf{Entailment} (0.98). \\
            \hspace*{2em} $\to$ \textbf{Reward $Q(s,a)$:} \textcolor{border_success}{\textbf{+1.0}} \textit{(Proceed)}

            \vspace{2cm}
            \begin{tcolorbox}[colback=white, colframe=border_success, sharp corners]
                \centering
                \textbf{Final Prediction:} \\
                \textcolor{border_success}{\textbf{The Weekly Standard (1995)}} \\
                \ding{51} \textit{Factually Grounded}
            \end{tcolorbox}
        \end{tcolorbox}
    \end{minipage}

    \caption{\textbf{Comprehensive Trace of Failure vs. Correction.} 
    \textbf{Left:} The Baseline fails due to \textit{information crowding}. High-similarity redundant documents about Entity A fill the context window, cutting off Entity B. The model then hallucinates to fill the gap. 
    \textbf{Right:} Our approach employs \textbf{MMKP} (Maximal Marginal Relevance based Knapsack Problem) to filter semantic duplicates, ensuring both entities are present. Subsequently, the \textbf{MCTS} (Monte Carlo Tree Search) planner explores reasoning paths. It actively penalizes the hallucinated branch (Branch 1) via NLI verification and rewards the factually consistent branch (Branch 2).}
    \label{fig:comprehensive_trace}
\end{figure*}

\begin{figure*}[t]
    \centering
    \small
    
    \begin{tcolorbox}[colback=bg_query, colframe=gray!40, title={\textsc{Input Query \& Complexity Analysis}}, coltitle=black]
        \textbf{Query:} \textit{"Who had a longer tenure as CEO of the company that acquired DeepMind: Eric Schmidt or Larry Page?"}
        \par\noindent\rule{\textwidth}{0.4pt}
        \textbf{Reasoning Requirement:} 
        1. \textbf{Entity Bridging:} Identify Parent Company (DeepMind $\to$ Google). 
        2. \textbf{Attribute Retrieval:} Find CEO Tenure(Schmidt) AND CEO Tenure(Page).
        3. \textbf{Numerical Comparison:} Calculate $\Delta t$.
        \hfill \textit{Hardness: High (Distractor Noise)}
    \end{tcolorbox}

    \vspace{2mm}

    \begin{minipage}[t]{0.49\textwidth}
        \begin{tcolorbox}[
            colback=bg_fail, colframe=border_fail, 
            title={\textcolor{black}{\faSearch\ Baseline: Standard RAG (Failure Trace)}}, 
            coltitle=border_fail,
            equal height group=ragtrace
        ]
            
            \textbf{\textsc{Step 1: Dense Retrieval (Top-K)}} \\
            \textit{Retriever matches "DeepMind" to high-ranking news chunks.}
            
            \begin{itemize}
                \setlength\itemsep{0em}
                \item \textbf{[Rank 1] Doc A:} "Google acquires AI startup DeepMind for \$500M in 2014..." \\
                \footnotesize \textcolor{gray}{$\to$ Sim: 0.91 (Relevant Bridge)} \normalsize
                
                \item \textbf{[Rank 2] Doc B:} "DeepMind's AlphaGo defeats Lee Sedol in historic match..." \\
                \footnotesize \textcolor{gray}{$\to$ Sim: 0.89 (Distractor - High Semantic Overlap)} \normalsize
                
                \item \textbf{[Rank 3] Doc C:} "Demis Hassabis discusses DeepMind's AlphaZero algorithm..." \\
                \footnotesize \textcolor{gray}{$\to$ Sim: 0.87 (Distractor - Redundant Topic)} \normalsize
            \end{itemize}
            
            \vspace{1mm}
            \textbf{\textcolor{red}{CRITICAL ERROR:}} \textit{Distractor Crowding.} High-scoring documents about DeepMind's AI achievements fill the context. \textit{Evidence for Schmidt's tenure is pushed to Rank 7 and truncated.}

            \par\noindent\rule{\textwidth}{0.4pt}

            \textbf{\textsc{Step 2: CoT Reasoning (Hallucination)}} \\
            \textit{LLM Input: Doc A, Doc B, Doc C.}
            
            \begin{itemize}
                \item \textbf{Thought 1:} "DeepMind was acquired by Google (Doc A)."
                \item \textbf{Thought 2:} "Need tenures for Eric Schmidt and Larry Page."
                \item \textbf{Thought 3 (Missing Info):} "Context discusses AlphaGo, not CEO dates."
                \item \textbf{Thought 4 (Heuristic Bias):} "Larry Page is the co-founder of Google. Founders usually serve the longest terms. Eric Schmidt was hired later."
            \end{itemize}

            \vspace{1mm}
            \begin{tcolorbox}[colback=white, colframe=border_fail, sharp corners]
                \centering
                \textbf{Final Prediction:} \\
                \textcolor{border_fail}{\textbf{Larry Page (Founder Heuristic)}} \\
                \ding{55} \textit{Factually Incorrect (Page: $\approx$4 yrs vs Schmidt: 10 yrs)}
            \end{tcolorbox}
        \end{tcolorbox}
    \end{minipage}
    \hfill
    \begin{minipage}[t]{0.49\textwidth}
        \begin{tcolorbox}[
            colback=bg_success, colframe=border_success, 
            title={\textcolor{black}{\faFilter\ Ours: Self-Correcting RAG (Success Trace)}}, 
            coltitle=border_success,
            equal height group=ragtrace
        ]
            
            \textbf{\textsc{Phase 1: MMKP Context Selector}} \\
            \textit{Goal: Penalize Semantic Redundancy ($C_{red}$).}
            
            \begin{itemize}
                \setlength\itemsep{0em}
                \item \textbf{Cluster 1 (DeepMind News):} \{Doc B (AlphaGo), Doc C (AlphaZero)\}
                \item \textbf{Cluster 2 (Corporate Structure):} \{Doc A (Acquisition), Doc D (Schmidt Tenure), Doc E (Page Tenure)\}
                \item \textbf{Action:} Detect high redundancy in Cluster 1. \textbf{Prune Doc B \& C}. Allocate budget to Cluster 2.
            \end{itemize}
            \textbf{Optimized Context:} \{Doc A, Doc D ("Schmidt CEO 2001-2011"), Doc E ("Page CEO 2011-2015")\}
             \vspace{0.2cm} 
            \par\noindent\rule{\textwidth}{0.4pt}

            \textbf{\textsc{Phase 2: MCTS Planner Reasoning}} \\
            \textit{Node 0: Root State (Question)}
            
            \vspace{0.4cm}
            \textbf{$\hookrightarrow$ Branch 1: Heuristic Assumption (Fail)} \\
            \hspace*{1em} \textit{Hypothesis:} "Larry Page served longer." \\
            \hspace*{1em} $\to$ \textbf{Verification (NLI):} Context (Schmidt: 10 yrs, Page: 4 yrs) vs Hypothesis. \\
            \hspace*{1em} $\to$ \textbf{Result:} \textbf{Contradiction} (0.95). \\
            \hspace*{1em} $\to$ \textbf{Reward:} \textcolor{red}{\textbf{-1.0}} \textit{(Prune)}

            \vspace{0.4cm}
            \textbf{$\hookrightarrow$ Branch 2: Calculation (Success)} \\
            \hspace*{1em} \textit{Step 1:} Schmidt: $2011 - 2001 = 10$ years. \\
            \hspace*{1em} \textit{Step 2:} Page: $2015 - 2011 = 4$ years. \\
            \hspace*{1em} \textit{Hypothesis:} "Eric Schmidt (10 years) > Larry Page." \\
            \hspace*{1em} $\to$ \textbf{Verification (NLI):} \textbf{Entailment} (0.99). \\
            \hspace*{1em} $\to$ \textbf{Reward:} \textcolor{border_success}{\textbf{+1.0}} \textit{(Proceed)}

            \vspace{3.3cm}
            \begin{tcolorbox}[colback=white, colframe=border_success, sharp corners]
                \centering
                \textbf{Final Prediction:} \\
                \textcolor{border_success}{\textbf{Eric Schmidt (10 years)}} \\
                \ding{51} \textit{Factually Grounded Calculation}
            \end{tcolorbox}
        \end{tcolorbox}
    \end{minipage}

    \caption{\textbf{Analysis of Distractor Filtering and Numerical Verification.} 
    \textbf{Left:} The Baseline fails due to \textit{Distractor Crowding}. Popular documents about DeepMind's AI achievements (AlphaGo) overwhelm the context window, displacing the necessary CEO tenure dates. 
    \textbf{Right:} The \textbf{MMKP Selector} identifies the "AI Achievement" documents as semantically redundant and removes them. This preserves space for documents containing specific tenure dates. The \textbf{MCTS Planner} then rejects the heuristic bias ("Founders serve longer") via NLI verification, ensuring the final answer is derived from arithmetic comparison of the retrieved dates.}
    \label{fig:deepmind_trace}
\end{figure*}

\end{document}